\pdfoutput=1

\documentclass[11pt]{article}
\usepackage{CJKutf8}
\usepackage[final]{acl}

\usepackage{times}
\usepackage{latexsym}
\usepackage{makecell}
\usepackage{booktabs}
\usepackage{listings}

\usepackage[utf8]{inputenc}

\usepackage{microtype}

\usepackage{inconsolata}
\usepackage[export]{adjustbox} %

\usepackage{graphicx}
\usepackage[utf8]{inputenc}  %
\usepackage{multirow}
\usepackage{colortbl} %
\usepackage{xcolor}   %
\usepackage{fancyvrb}
\usepackage{anyfontsize}
\usepackage{placeins}
\usepackage{marvosym}

\title{SmartBench: Is Your LLM Truly a Good Chinese Smartphone Assistant?}

\author{Xudong Lu\textsuperscript{*1,2}, Haohao Gao\textsuperscript{*1}, Renshou Wu\textsuperscript{*$\dagger$1}, Shuai Ren\textsuperscript{1}, Xiaoxin Chen\textsuperscript{1}, \\ \textbf{Hongsheng Li\textsuperscript{2}$^{\textrm{\Letter}}$, Fangyuan Li\textsuperscript{1}$^{\textrm{\Letter}}$}\\
$^1$vivo AI Lab\quad $^2$CUHK MMLab\\
$^*$Equal contribution $^{\textrm{\Letter}}$Corresponding author $^\dagger$Project lead\\
\texttt{\{luxudong@link,hsli@ee\}.cuhk.edu.hk}\quad \texttt{\{gaohaohao,wurenshou,lifangyuan\}@vivo.com}
}

\begin{document}
\maketitle
\begin{abstract}
Large Language Models (LLMs) have become integral to daily life, especially advancing as intelligent assistants through on-device deployment on smartphones. However, existing LLM evaluation benchmarks predominantly focus on objective tasks like mathematics and coding in English, which do not necessarily reflect the practical use cases of on-device LLMs in real-world mobile scenarios, especially for Chinese users. To address these gaps, we introduce \textbf{SmartBench}, the first benchmark designed to evaluate the capabilities of on-device LLMs in Chinese mobile contexts. We analyze functionalities provided by representative smartphone manufacturers and divide them into five categories: text summarization, text Q\&A, information extraction, content creation, and notification management, further detailed into 20 specific tasks. For each task, we construct high-quality datasets comprising 50 to 200 question-answer pairs that reflect everyday mobile interactions, and we develop automated evaluation criteria tailored for these tasks. We conduct comprehensive evaluations of on-device LLMs and MLLMs using SmartBench and also assess their performance after quantized deployment on real smartphone NPUs. Our contributions provide a standardized framework for evaluating on-device LLMs in Chinese, promoting further development and optimization in this critical area. Code and data will be available at \url{https://github.com/vivo-ai-lab/SmartBench}.
\end{abstract}

\section{Introduction}

Large Language Models (LLMs) have significantly transformed everyday life in recent years by serving as intelligent, context-aware assistants~\cite{openai2024gpt4o,team2024gemini,anthropic2023claude3,anil2023gemini,lu2024deepseekvl,jiang2024mixtral,abdin2024phi,guo2025deepseek}. To further enhance the capabilities of LLMs in serving human needs, various academic research and engineering efforts have focused on deploying smaller LLMs on edge devices, such as smartphones~\cite{xue2024powerinfer,yao2024minicpm,chu2023mobilevlm,chu2024mobilevlm,lu2024bluelm}. As companions in our daily lives, smartphones serve as crucial platforms for people to experience the capabilities of on-device LLMs. The local deployment of LLMs on end-side smartphones eliminates the need for a network connection, which not only broadens the scope of possible application scenarios but also enhances user privacy by keeping sensitive data processing on the device~\cite{qu2024mobile,ding2024enhancing}.

The current trend in smartphone technology shows that major manufacturers are increasingly adopting on-device LLMs~\cite{ashkboos2024computational}, integrating advanced AI capabilities into their devices. Industry leaders such as Apple with OpenELM~\cite{mehta2024openelm}, HUAWEI's Pangu E~\cite{zeng2021pangu},  Xiaomi’s MiLM~\cite{xiaomi_milm2_evolution}, and vivo's BlueLM-3B~\cite{lu2024bluelm} have demonstrated significant progress in this domain. These on-device LLMs support various real-time tasks~\cite{wu2024first}, offering users seamless and responsive AI-powered mobile interactions~\cite{xu2024device}.
\begin{table*}[t]
\vspace{-3em}
\renewcommand{\arraystretch}{0.95}
\resizebox{\textwidth}{!}{%
\begin{tabular}{|llllllllll|}
\midrule
\multicolumn{4}{|c|}{\textbf{Text   Summarization}} & \multicolumn{3}{c|}{\textbf{Text Q\&A}} & \multicolumn{3}{c|}{\textbf{Information Extraction}} \\ \midrule
\multicolumn{1}{|l|}{\begin{tabular}[c]{@{}l@{}}Document\\ Summ\\\begin{CJK}{UTF8}{gbsn}文档摘要\end{CJK}
\end{tabular}} &
  \multicolumn{1}{l|}{\begin{tabular}[c]{@{}l@{}}Call\\ Summ\\\begin{CJK}{UTF8}{gbsn}通话摘要\end{CJK}\end{tabular}} &
  \multicolumn{1}{l|}{\begin{tabular}[c]{@{}l@{}}Recording\\ Summ\\\begin{CJK}{UTF8}{gbsn}录音摘要\end{CJK}\end{tabular}} &
  \multicolumn{1}{l|}{\begin{tabular}[c]{@{}l@{}}Meeting\\ Summ\\\begin{CJK}{UTF8}{gbsn}会议摘要\end{CJK}\end{tabular}} &
  \multicolumn{1}{l|}{\begin{tabular}[c]{@{}l@{}}Document\\ Q\&A\\\begin{CJK}{UTF8}{gbsn}文档问答\end{CJK}\end{tabular}} &
  \multicolumn{1}{l|}{\begin{tabular}[c]{@{}l@{}}Retrieval\\ Q\&A\\\begin{CJK}{UTF8}{gbsn}检索问答\end{CJK}\end{tabular}} &
  \multicolumn{1}{l|}{\begin{tabular}[c]{@{}l@{}}Personal\\ Q\&A\\\begin{CJK}{UTF8}{gbsn}个人问答\end{CJK}\end{tabular}} &
  \multicolumn{1}{l|}{\begin{tabular}[c]{@{}l@{}}Entity\\ Extraction\\\begin{CJK}{UTF8}{gbsn}实体抽取\end{CJK}\end{tabular}} &
  \multicolumn{1}{l|}{\begin{tabular}[c]{@{}l@{}}Relation\\ Extraction\\\begin{CJK}{UTF8}{gbsn}关系抽取\end{CJK}\end{tabular}} &
  \begin{tabular}[c]{@{}l@{}}Event\\ Extraction\\\begin{CJK}{UTF8}{gbsn}事件抽取\end{CJK}\end{tabular} \\ \midrule
\multicolumn{8}{|c|}{\textbf{Content Creation}}                                                    & \multicolumn{2}{c|}{\textbf{Notification Management}} \\ \midrule
\multicolumn{1}{|l|}{\begin{tabular}[c]{@{}l@{}}Text\\ Polishing\\\begin{CJK}{UTF8}{gbsn}文本润色\end{CJK}\end{tabular}} &
  \multicolumn{1}{l|}{\begin{tabular}[c]{@{}l@{}}Text\\ Continuation\\\begin{CJK}{UTF8}{gbsn}文本续写\end{CJK}\end{tabular}} &
  \multicolumn{1}{l|}{\begin{tabular}[c]{@{}l@{}}Text\\ Abbreviation\\\begin{CJK}{UTF8}{gbsn}文本缩写\end{CJK}\end{tabular}} &
  \multicolumn{1}{l|}{\begin{tabular}[c]{@{}l@{}}Text\\ Expansion\\\begin{CJK}{UTF8}{gbsn}文本扩写\end{CJK}\end{tabular}} &
  \multicolumn{1}{l|}{\begin{tabular}[c]{@{}l@{}}Text\\ Creation\\\begin{CJK}{UTF8}{gbsn}文本创作\end{CJK}\end{tabular}} &
  \multicolumn{1}{l|}{\begin{tabular}[c]{@{}l@{}}Text\\ Formatting\\\begin{CJK}{UTF8}{gbsn}文本排版\end{CJK}\end{tabular}} &
  \multicolumn{1}{l|}{\begin{tabular}[c]{@{}l@{}}Instant\\ Reply\\\begin{CJK}{UTF8}{gbsn}即时回复\end{CJK}\end{tabular}} &
  \multicolumn{1}{l|}{\begin{tabular}[c]{@{}l@{}}Text\\ Correction\\\begin{CJK}{UTF8}{gbsn}文本纠错\end{CJK}\end{tabular}} &
  \multicolumn{1}{l|}{\begin{tabular}[c]{@{}l@{}}Notification\\ Sorting\\\begin{CJK}{UTF8}{gbsn}通知排序\end{CJK}\end{tabular}} &
  \begin{tabular}[c]{@{}l@{}}Message\\ Summ\\\begin{CJK}{UTF8}{gbsn}消息总结\end{CJK}\end{tabular} \\ \midrule
\end{tabular}%
}
\vspace{-0.5em}
\caption{We analyze the on-device LLM features currently released on mobile phones by major manufacturers, dividing them into 5 major categories with 20 tasks. Based on this, we propose SmartBench, the first (Chinese) benchmark for assessing the capabilities of on-device LLMs in mobile scenarios.}
\vspace{-1em}
\label{tab:category}
\end{table*}

However, we find that there are still notable gaps in the comprehensive evaluations for assessing the capabilities of on-device LLMs deployed on smartphones. Traditional LLM evaluations are typically categorized into two dimensions, i.e., \textbf{\textit{objective}} tasks and \textbf{\textit{subjective}} tasks. Objective tasks primarily focus on the assessment of knowledge, encompassing areas such as mathematical proficiency with benchmarks like GSM8K~\cite{cobbe2021training} and MATH~\cite{hendrycks2021measuring}, coding competence evaluated through HumanEval~\cite{chen2021evaluating}, and multitask accuracy measured by MMLU~\cite{hendrycks2020measuring}. Subjective tasks typically evaluate the model's ability to generate coherent, contextually appropriate, and human-like responses. These tasks often consider the model's creativity, fluency, adaptability to nuanced instructions, and alignment with user intent. Subjective evaluation datasets are often derived from user-constructed scenarios~\cite{liu2023alignbench}, curated human-chatbot conversations~\cite{lin2024wildbench}, and filtered interactions from platforms like Chatbot Arena~\cite{li2024crowdsourced,arenahard2024}. For on-device smartphone applications, the evaluation predominantly emphasizes subjective capabilities. Through our investigation, we identify the following critical gaps in existing subjective evaluation benchmarks:

\textbf{1) The scenario gap}: Current benchmarks emphasize tasks like mathematics and coding~\cite{alpaca_eval,lin2024wildbench,arenahard2024}, which are rarely handled by on-device LLMs in practical applications. Instead, on-device LLMs place greater emphasis on lightweight tasks such as text refinement, and notification processing.

\textbf{2) The language gap}: Mobile users who speak different languages often have varying living environments and language habits. Currently, most evaluation protocols for subjective tasks are all in English. As a market with over 1 billion smartphone users~\cite{statista2025china}, it is crucial to have an evaluation benchmark for LLMs deployed on Chinese-oriented smartphones.

To tackle these gaps, in this paper, starting from a functional investigation of on-device LLMs, we construct SmartBench, the first (Chinese) benchmark for evaluating the capabilities of on-device LLMs in mobile scenarios. Specifically, we analyze the on-device LLM functionalities provided by Apple, HUAWEI, OPPO, vivo, Xiaomi, and HONOR (up to December 2024), dividing them into five categories: text summarization, text Q\&A, information extraction, content creation, and notification management. Building on these functionalities, we further refine the 5 categories into 20 tasks, as outlined in Tab.~\ref{tab:category}. To evaluate each task, we construct 50 to 200 question-answer (QA) pairs per task that reflect everyday life scenarios by screening open-source datasets and generating additional pairs using manual collection or LLMs, resulting in a total of 2973 QA pairs. Evaluations of subjective tasks are commonly conducted using the LLM-as-a-Judge paradigm~\cite{zheng2023judging}. In SmartBench, we develop detailed automated evaluation criteria for each category/task. We further conduct comprehensive evaluations of multiple on-device LLMs and MLLMs on SmartBench and assess their performance after quantized deployment on the NPU of real smartphones.

Our contributions are summarized as follows:

\textbf{1)} We investigate the on-device LLM features offered by representative smartphone manufacturers, organizing them into 5 categories comprising 20 tasks. We then introduce SmartBench, the first Chinese benchmark designed to evaluate the capabilities of on-device LLMs in mobile scenarios,  featuring 2973 QA pairs.

\textbf{2)} For each task, we construct high-quality text QA pairs tailored to mobile usage scenarios by screening open-source datasets, manually collecting data, and synthesizing data using LLMs. Additionally, we develop high-quality automated evaluation methods for each category/task.

\textbf{3)} We evaluate the performance of representative end-side LLMs/MLLMs using SmartBench. Additionally, we assess the accuracy of quantized models running on real smartphone NPUs, which offers greater practical value.

\section{Related Works}\label{sec:relate}

\begin{figure*}[t]
    \centering
    \vspace{-3em}
    \includegraphics[
        width=0.98\linewidth,
        margin=0 0 0 0,  %
        clip                 %
    ]{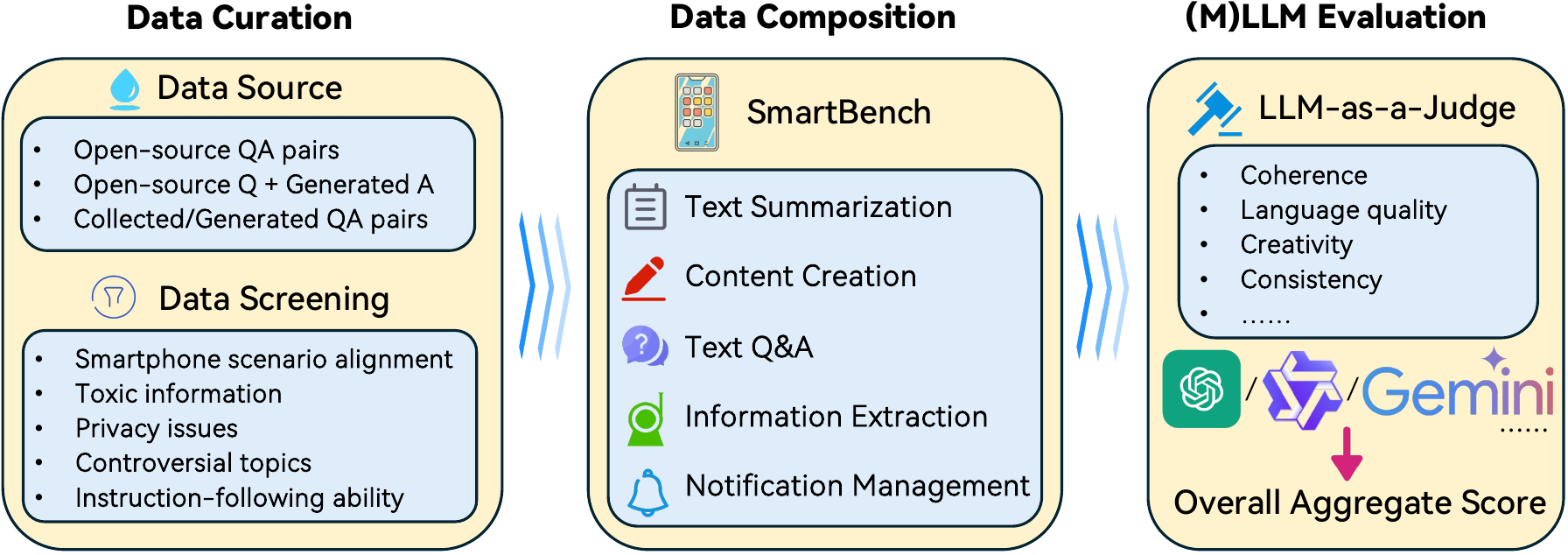}
    \caption{Overview of SmartBench, including data curation, data composition, and LLM-as-a-Judge evaluation.}
    \label{fig:overview}
    \vspace{-1em}
\end{figure*}

\subsection{Large Language Models on Edge Devices}

The deployment of LLMs on edge devices has garnered significant attention in recent years. In the academic community, there are currently numerous open-source LLMs and MLLMs, such as Qwen2.5 3B~\cite{yang2024qwen2}, InternVL 2.5 4B~\cite{chen2024expanding}, and MiniCPM 3.0 4B~\cite{hu2024minicpm}. Most of these models have between 3B and 4B parameters, making them well-suited for deployment on edge devices with limited computational capabilities. Besides, major smartphone manufacturers have also introduced their own LLMs, including Gemini Nano by Google, BlueLM by vivo, Magic LM by HONOR, OpenELM by Apple, and MiLM by Xiaomi~\cite{wu2024first}. These advancements pave the way for more efficient and powerful AI applications on edge devices.

\subsection{Benchmarks for Realworld Assistance}

How to comprehensively evaluate LLMs has long been a widely researched topic~\cite{chang2024survey}. The vast majority of benchmarks are designed to assess the knowledge capabilities of these models, including general knowledge~\cite{hendrycks2020measuring, wang2024mmlu, allenai_arc}, mathematics and science knowledge~\cite{cobbe2021training, hendrycks2021measuring, rein2023gpqa}, and coding ability~\cite{austin2021program, chen2021evaluating}, etc. Recently, there have been new datasets introduced to test the ability of models to handle real users' questions in the wild~\cite{liu2023alignbench,lin2024wildbench}. These datasets often consist of subjective questions that focus on the creativity and ability of models to follow instructions in real-world usage scenarios~\cite{arenahard2024,li2024crowdsourced,alpaca_eval}, providing a more direct reflection of user comfort and satisfaction during real-world usage. SmartBench is the first benchmark designed to evaluate the practical functionalities of LLMs deployed on smartphones.

\subsection{Chinese LLM Benchmarks}

With the rapid development of Chinese LLMs~\cite{sun2021ernie,2023internlm,guo2025deepseek}, specialized evaluation benchmarks have been established to assess their performance in understanding and generating content within a Chinese context. Prominent Chinese LLM benchmarks include CMRC~\cite{cui-emnlp2019-cmrc2018}, CLUE~\cite{xu2020clue}, SuperCLUE~\cite{xu2023superclue}, and C-Eval~\cite{huang2023ceval}, etc. Additionally, there are datasets like AlignBench~\cite{liu2023alignbench} designed for evaluating subjective tasks in Chinese. However, SmartBench distinguishes itself by focusing specifically on everyday mobile scenarios, offering a unique perspective on the practical functionalities of on-device LLMs in real-life smartphone usage.

\subsection{LLM Agent on Smartphones}

There is another type of task on mobile phones that helps solve real-world tasks, called mobile agents~\cite{wang2024mobile, zhang2023appagent, chai2024amex,rawles2024androidworld}. These tasks often involve executing multi-step commands on the phone based on user instructions~\cite{zhang2024android}. In contrast, Smartbench focuses on the functionality of on-device LLMs for handling common daily tasks in a single step, without planning action trajectories or calling external APIs.

\section{SmartBench}\label{sec:method}

In this section, we present a detailed description of the proposed SmartBench benchmark, specifically focusing on the scenario of smartphone deployment. We cover the data composition (Sec.~\ref{sec:data_comp}), data sources (Sec.~\ref{sec:data_source}), filtering criteria (Sec.~\ref{sec:data_screen}), and evaluation protocol (Sec.~\ref{sec:eval_pro}) used in the construction of the benchmark. The overview of SmartBench is illustrated in Fig.~\ref{fig:overview}.

\begin{figure*}
    \centering
    \vspace{-3em}
    \includegraphics[
        width=1.05\linewidth,
        margin=-1em 0 0 0,  %
        clip                 %
    ]{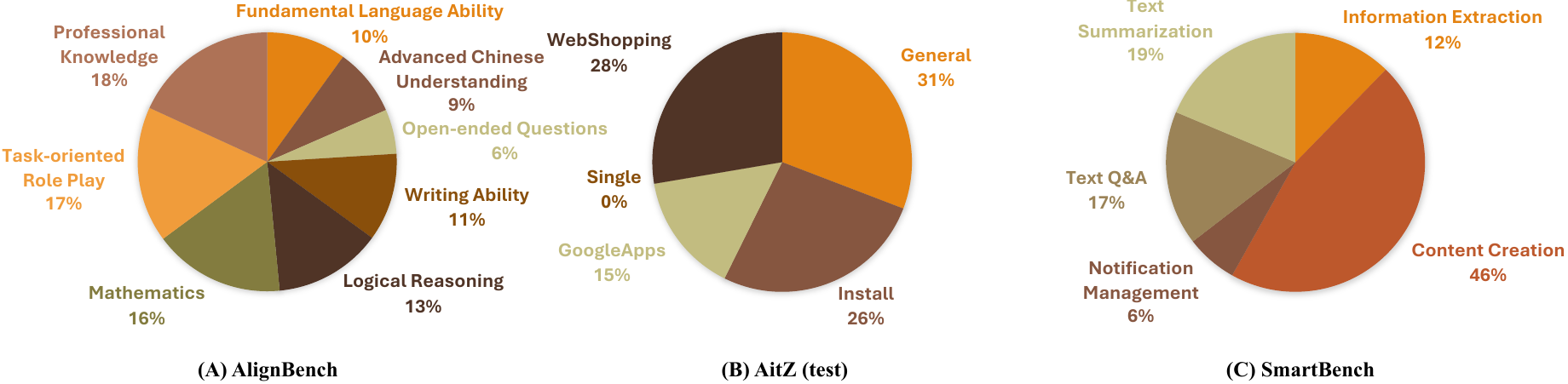}
    \caption{Data composition comparison between AlignBench~\cite{liu2023alignbench}, AitZ~\cite{zhang2024android} and SmartBench. AlignBench (zh) is a general benchmark designed for Chinese scenarios, and AitZ (en) is a mobile agent benchmark. SmartBench (zh) is specifically designed for evaluating end-side LLM functionality on smartphones.}
    \label{fig:data_comp}
    \vspace{-0.5em}
\end{figure*}

\subsection{Data Composition}\label{sec:data_comp}

We divide the on-device LLM features released by representative smartphone manufacturers into 5 categories, encompassing a total of 20 tasks.

\textbf{1) Text Summarization:} This category is focused on providing a concise summary of the text in one sentence and listing key information in bullet points. The benefit of this function is that it allows users to quickly grasp the main ideas and essential details without needing to read through lengthy content. We categorize the content into four scenarios. Document summarization primarily targets emails, scientific knowledge, and news reports. Call summarization focuses on conversations between two people. Recording summarization focuses on recordings that have significant background noise. Meeting summarization specifically refers to the summarization of meetings.

\textbf{2) Content Creation:} This category highlights the functionality of creating content on mobile devices, enabling users to effortlessly share their creations on social media platforms such as Weibo, WeChat Moments, and RedNote (Xiaohongshu). With the widespread use of smartphones, mobile content creation has become increasingly accessible and convenient. We focus on the commonly utilized functions for content creation, i.e., polishing, continuation, abbreviation, expansion, (automatic) creation, formatting, and correction. Additionally, on-device LLMs are employed to refine users' message replies; therefore, we also incorporate tests for the instant reply functionality.

\begin{table}[t]\scriptsize
\renewcommand\arraystretch{0.6}
\resizebox{\columnwidth}{!}{%
\begin{tabular}{lc|cc}
\midrule
\textbf{Category} & \textbf{Count} & \textbf{Input} & \textbf{Target} \\ \midrule
Text Summarization & 550 & 1890 & 244 \\ \midrule
Content Creation & 1377 & 210 & 143 \\ \midrule
Text Q\&A & 495 & 930 & 115 \\ \midrule
Information Extraction & 362 & 682 & 74 \\ \midrule
Notification Management & 189 & 376 & 101 \\ \midrule
\end{tabular}%
}
\vspace{-0.5em}
\caption{We present the number of QA pairs for each category in SmartBench. For each category, we also provide the average input (query) length and the average target (reference answer) length of all QA pairs.}
\label{tab:count}
\vspace{-2em}
\end{table}

\textbf{3) Text Q\&A:} This feature allows users to quickly obtain information or answer questions through simple text inputs. We categorize it into three scenarios: Document Q\&A, where a specific document is provided and questions are answered based on it; Retrieval Q\&A, where answers are summarized based on multiple relevant retrieval contents and questions; and Personal Q\&A, where information from synthesized personal records (such as memos or personal notes) is used.

\textbf{4) Information Extraction}: This category involves automatically identifying and extracting specific data from text inputs, such as names, dates, addresses, or other relevant information. The information extraction functionality on mobile phones is primarily divided into three aspects: Entity Extraction, which involves identifying and extracting specific entities from text, such as names, locations, dates, etc.; Relation Extraction, which analyzes and extracts relationships between entities, such as ``someone works at a certain company''; and Event Extraction, which identifies specific events and their related elements from text, such as time, location, and participants. These functionalities collectively contribute to intelligent applications, such as automatic summarization, smart search, and personalized recommendations.

\begin{figure}[t]
    \centering
    \includegraphics[width=\linewidth]{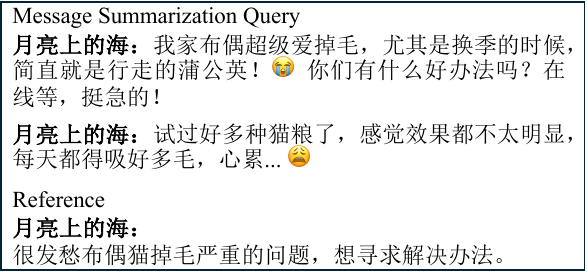}
    \caption{Example of the Message Summarization task in SmartBench (English translated version in Fig.~\ref{fig:en_case_example}).}
    \label{fig:case_cat}
    \vspace{-1.5em}
\end{figure}

\begin{figure*}[t]
\vspace{-3em}
\includegraphics[width=\textwidth]{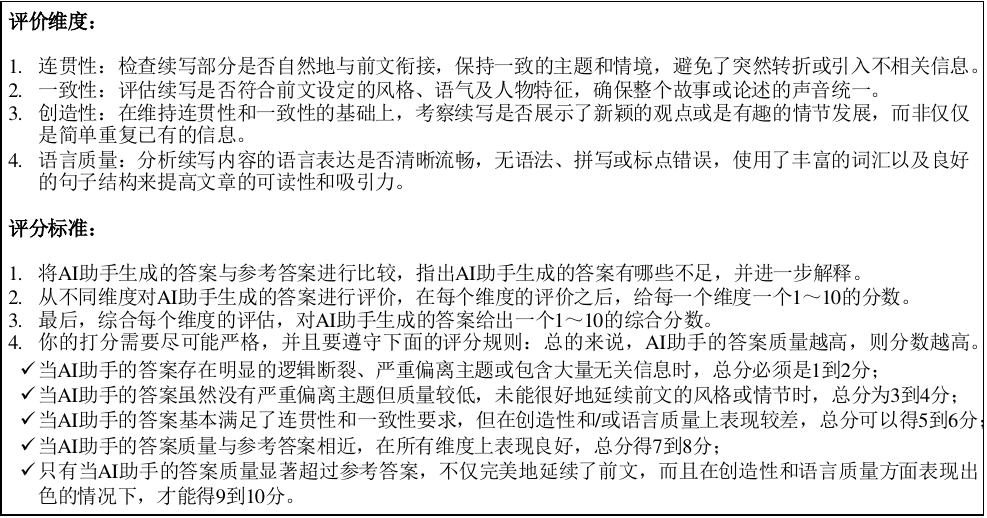}
\vspace{-1.5em}
\caption{Evaluation Dimension \& Scoring Standard for the text continuation task (English version in Fig.~\ref{fig:en_eval_prompt}).}
\vspace{-1em}
\label{fig:score_prompt}
\end{figure*}

\begin{table*}[t]\huge
\vspace{-2em}
\renewcommand\arraystretch{1}
\resizebox{\textwidth}{!}{%
\begin{tabular}{clc|ccccc}
\midrule
\multicolumn{1}{c}{\textbf{Category}} & \textbf{Task} & \textbf{GPT-4o} & \textbf{BlueLM-3B} & \textbf{InternVL2.5-4B} & \textbf{MiniCPM3-4B} & \textbf{Qwen2.5-3B} & \textbf{Qwen2-VL-2B} \\ \midrule
\multicolumn{1}{c}{\multirow{4}{*}{\begin{tabular}[c]{@{}c@{}}Text\\ Summarization\end{tabular}}} & Document Summ & 7.05 & \textbf{7.56} & 6.89 & 7.40 & 7.21 & 4.37 \\ \cmidrule{2-8} 
\multicolumn{1}{c}{} & Call Summ & 7.03 & \textbf{7.22} & 5.43 & 6.88 & 6.35 & 3.48 \\ \cmidrule{2-8} 
\multicolumn{1}{c}{} & Recording Summ & 7.78& \textbf{7.63} & 6.38 & 7.45 & 7.07 & 4.17 \\ \cmidrule{2-8} 
\multicolumn{1}{c}{} & Meeting Summ & 7.73& \textbf{7.09} & 6.23 & 6.98 & 6.67 & 3.75 \\ \midrule
\multicolumn{1}{c}{\multirow{3}{*}{Text Q\&A}} & Document Q\&A & 8.83& \textbf{9.37} & 9.36 & 8.39 & 9.34 & 9.15 \\ \cmidrule{2-8} 
\multicolumn{1}{c}{} & Retrieval Q\&A & 6.91 & 5.89 & 5.81 & \textbf{6.76} & 6.25 & 4.77 \\ \cmidrule{2-8} 
\multicolumn{1}{c}{} & Personal Q\&A & 9.78 & 9.36 & 8.89 & 8.87 & \textbf{9.39} & 8.83 \\ \midrule
\multicolumn{1}{c}{\multirow{8}{*}{\begin{tabular}[c]{@{}c@{}}Content\\Creation\end{tabular}}} & Text Polishing & 7.49& \textbf{7.55} & 6.17 & 7.53 & 7.42 & 6.19 \\ \cmidrule{2-8} 
\multicolumn{1}{c}{} & Text Continuation & 7.63 & 7.45 & 6.89 & 7.52 & \textbf{7.72} & 5.96 \\ \cmidrule{2-8} 
\multicolumn{1}{c}{} & Text Abbreviation & 7.49 & 8.23 & 7.43 & 8.17 & \textbf{8.51} & 7.51 \\ \cmidrule{2-8} 
\multicolumn{1}{c}{} & Text Expansion & 8.42 & 7.44 & 8.04 & \textbf{8.74} & 8.07 & 6.04 \\ \cmidrule{2-8} 
\multicolumn{1}{c}{} & Text Creation & 7.55 & \textbf{6.93} & 6.16 & 6.89 & 6.68 & 5.26 \\ \cmidrule{2-8} 
\multicolumn{1}{c}{} & Text Formatting & 6.15 & 6.03 & 5.10 & \textbf{6.80} & 3.69 & 1.20 \\ \cmidrule{2-8} 
\multicolumn{1}{c}{} & Instant Reply & 7.62 & \textbf{6.70} & 5.90 & 6.28 & {6.44} & 3.14 \\ \cmidrule{2-8} 
\multicolumn{1}{c}{} & Text Correction & 7.03 & \textbf{3.69} & 2.46 & 3.24 & 2.38 & 1.17 \\ \midrule
\multicolumn{1}{c}{\multirow{3}{*}{\begin{tabular}[c]{@{}c@{}}Information\\Extraction\end{tabular}}} & Entity Extraction & 8.36 & 7.82 & \textbf{8.13} & 7.58 & 6.35 & 5.00 \\ \cmidrule{2-8} 
\multicolumn{1}{c}{} & Relation Extraction & 3.54 & \textbf{5.55} & 3.58 & 4.15 & 3.54 & 3.04 \\ \cmidrule{2-8} 
\multicolumn{1}{c}{} & Event Extraction & 7.86 & 6.79 & \textbf{7.09} & 6.20 & 6.75 & 4.66 \\ \midrule
\multicolumn{1}{c}{\multirow{2}{*}{\begin{tabular}[c]{@{}c@{}}Notification\\Management\end{tabular}}} & Message Summ &  7.24 & 7.45 & 7.29 & \textbf{8.08} & 7.86 & 5.90 \\ \cmidrule{2-8} 
\multicolumn{1}{c}{} & Notification Sorting & 5.97 & 4.78 & 4.19 & \textbf{4.85} & 4.51 & 2.14 \\ \midrule
\rowcolor{gray!20}
\multicolumn{2}{c}{AVG} & 7.37 & \textbf{7.03} & 6.37 & 6.94 & 6.61 & 4.79 \\ \midrule
\end{tabular}%
}
\caption{Evaluation results using GPT-4 Turbo (\texttt{gpt-4-turbo-04-09}) as the judge LLM. We compare BlueLM-3B, InternVL2.5-4B, MiniCPM3-4B, Qwen2.5-3B, and Qwen2-VL-2B (on-device models) on the whole SmartBench in BF16 precision with GPT-4o (on-cloud model) for reference. The scores assessed by Qwen-Max as the judge LLM are also provided in Tab.~\ref{tab:fp_qwen} in the Appendix.}
\vspace{-0.5em}
\label{tab:fp}
\end{table*}

\textbf{5) Notification Management:} Effective notification management on smartphones is essential to minimize distractions, enhance productivity, and ensure timely access to important information. Currently, LLMs deployed on smartphones primarily support two functions: Notification Sorting, which organizes and prioritizes notifications based on degree of urgency or chronological order; and Message Summarization, which condenses lengthy notifications or messages into concise summaries for quick understanding. By intelligently sorting and summarizing information, smartphones equipped with such features can significantly improve efficiency and reduce cognitive overload in our increasingly connected world.

To facilitate a clear comparison between SmartBench and other LLM benchmarks, we analyze their data composition. For the Chinese benchmark, we compare with AlignBench~\cite{liu2023alignbench}, while we select AitZ~\cite{zhang2024android} for the mobile agent benchmark, as illustrated in Fig.~\ref{fig:data_comp}. As shown, AlignBench serves as a more general benchmark for evaluating Chinese LLMs, AitZ focuses more on automated operations on mobile devices, while SmartBench emphasizes common on-device LLM functionalities. Additionally, we provide the number of QA pairs for each category in SmartBench in Tab.~\ref{tab:count}, along with the average input (query) length and the average target (reference answer) length for each category. Furthermore, to better illustrate the essence of SmartBench for mobile scenarios, we offer an example of the Message Summarization task in Fig.~\ref{fig:case_cat}.

\subsection{Data Source}\label{sec:data_source}

The data for SmartBench is primarily derived from three sources. 1) We screen open-source datasets to select QA pairs that align with smartphone application scenarios. 2) For datasets that provide contextual information but lack appropriate questions and answers, we utilize advanced LLMs, e.g., Qwen-Max~\cite{yang2024qwen2technicalreport}, Gemini Pro~\cite{reid2024gemini}, to generate corresponding answers for each task. 3) For the lack of open-source data in certain categories, we employ human collection and LLMs to generate QA pairs, followed by manual screening and editing to curate high-quality data.

For Text Summarization, we primarily use content from open-source datasets. For document data, we utilize the dataset from~\cite{bright_xu_2019_3402023}, which comprises a substantial Chinese corpus including content from Wikipedia, news reports, etc. For summarizing calls, recordings, and meetings, we draw data from Alimeeting4MUG~\cite{zhang2023mug}, LCCC~\cite{wang2020chinese}, VCSum~\cite{wu2023vcsum}, WenetSpeech~\cite{zhang2022wenetspeech}, etc. Speech content is converted to text transcriptions using speech-to-text converters in our benchmark, and the reference summaries for the summarization tasks are generated by Qwen-Max.

For Content Creation, we leverage QA pairs from CSCD-NS~\cite{hu2024cscd} for text correction. For other tasks, e.g., polishing, abbreviation, expansion, etc, we manually collect and design examples, and then use Gemini Pro and Qwen-Max to generate QA pairs tailored to meet the requirements of daily mobile usage scenarios.

For text Q\&A, we select document Q\&A pairs from the CMRC~\cite{cui-emnlp2019-cmrc2018} dataset. For retrieval-based Q\&A, the textual sources are from DuReader 2.0~\cite{he2017dureader}, and the answers are generated by Qwen-Max. For personal Q\&A, we design human examples and construct QA pairs (e.g., memos or personal notes) using Qwen-Max.

For Information Extraction, we source textual data for entity extraction from MSRA~\cite{levow-2006-third}, OntoNotes Release 4.0~\cite{weischedel2011ontonotes}, and Weibo~\cite{peng2016improving}. We use Qwen-Max to generate the corresponding answers. For relation and event extraction, we manually collect example data and generate textual information using Gemini Pro, then produce the corresponding answers with GPT-4 Turbo.

For Notification Management, we find that there is currently no suitable open-source data available for the smartphone platform. Therefore, we create human-designed examples and then use Gemini Pro to generate QA pairs for both notification sorting and message summarization.

\subsection{Data Screening}\label{sec:data_screen}

After initially collecting all the data for each task, we implement a rigorous screening process involving six domain experts with over five years of mobile AI experience. These specialists evaluate the dataset through dual-layer verification, primarily focusing on five core criteria: alignment with real-world smartphone interaction scenarios, detection of toxic or harmful information, identification of potential privacy leakage risks, flagging of socially controversial or polarizing topics, and comprehensive assessment of instruction-following capabilities of the reference answers.

To be specific, we implement a five-point scoring system (1–5) for each criterion, where 1 = Unacceptable, 2 = Poor, 3 = Fair, 4 = Good, and 5 = Very Good. Human experts score each item across all criteria, and we retain only those with an average score of $\geq$ 3.5, reducing the original 30k dataset to roughly 3k high-quality entries. For items scoring between 3.5 and 4, we perform manual re-labeling to guarantee accuracy and alignment with human judgment. All items are further refined and subjected to dual-layer verification, ensuring rigorous quality control throughout the dataset.

\begin{table*}[t]\small
\vspace{-3em}
\renewcommand\arraystretch{0.7}
\resizebox{\textwidth}{!}{%
\begin{tabular}{clcccccc}
\midrule
\multicolumn{1}{c}{\textbf{Category}} & \multicolumn{1}{c}{\textbf{Task}} & \multicolumn{3}{c}{\textbf{BlueLM-3B}} & \multicolumn{3}{c}{\textbf{Qwen2.5-3B}} \\ \midrule
\multicolumn{2}{c}{\textbf{Precision}} & \multicolumn{1}{c}{\textbf{BF16}} & \multicolumn{1}{c}{\textbf{INT4}} & \textbf{Retention (\%)} & \multicolumn{1}{c}{\textbf{BF16}} & \multicolumn{1}{c}{\textbf{INT4}} & \textbf{Retention (\%)} \\ \midrule
\multicolumn{1}{c}{\multirow{4}{*}{Text   Summarization}} & Document Summ & \multicolumn{1}{c}{7.22} & \multicolumn{1}{c}{4.98} & 68.92 & \multicolumn{1}{c}{6.89} & \multicolumn{1}{c}{4.44} & 64.52 \\ \cmidrule{2-8} 
\multicolumn{1}{c}{} & Call Summ & \multicolumn{1}{c}{7.00} & \multicolumn{1}{c}{6.77} & 96.73 & \multicolumn{1}{c}{6.86} & \multicolumn{1}{c}{6.29} & 91.67 \\ \cmidrule{2-8} 
\multicolumn{1}{c}{} & Recording Summ & \multicolumn{1}{c}{7.15} & \multicolumn{1}{c}{6.53} & 91.24 & \multicolumn{1}{c}{6.94} & \multicolumn{1}{c}{5.63} & 81.12 \\ \cmidrule{2-8} 
\multicolumn{1}{c}{} & Meeting Summ & \multicolumn{1}{c}{6.85} & \multicolumn{1}{c}{5.25} & 76.65 & \multicolumn{1}{c}{6.67} & \multicolumn{1}{c}{4.84} & 72.61 \\ \midrule
\multicolumn{1}{c}{\multirow{3}{*}{Text Q\&A}} & Document Q\&A & \multicolumn{1}{c}{9.77} & \multicolumn{1}{c}{9.54} & 97.64 & \multicolumn{1}{c}{9.77} & \multicolumn{1}{c}{9.38} & 96.06 \\ \cmidrule{2-8} 
\multicolumn{1}{c}{} & Retrieval Q\&A & \multicolumn{1}{c}{6.13} & \multicolumn{1}{c}{5.38} & 87.76 & \multicolumn{1}{c}{6.38} & \multicolumn{1}{c}{5.88} & 92.16 \\ \cmidrule{2-8} 
\multicolumn{1}{c}{} & Personal Q\&A & \multicolumn{1}{c}{8.71} & \multicolumn{1}{c}{7.58} & 87.04 & \multicolumn{1}{c}{9.29} & \multicolumn{1}{c}{9.15} & 98.54 \\ \midrule
\multicolumn{1}{c}{\multirow{8}{*}{Content Creation}} & Text Polishing & \multicolumn{1}{c}{7.57} & \multicolumn{1}{c}{7.18} & 94.81 & \multicolumn{1}{c}{7.54} & \multicolumn{1}{c}{7.07} & 93.84 \\ \cmidrule{2-8} 
\multicolumn{1}{c}{} & Text Continuation & \multicolumn{1}{c}{7.50} & \multicolumn{1}{c}{7.13} & 95.06 & \multicolumn{1}{c}{7.70} & \multicolumn{1}{c}{7.27} & 94.37 \\ \cmidrule{2-8} 
\multicolumn{1}{c}{} & Text Abbreviation & \multicolumn{1}{c}{7.81} & \multicolumn{1}{c}{7.06} & 90.50 & \multicolumn{1}{c}{8.23} & \multicolumn{1}{c}{7.27} & 88.33 \\ \cmidrule{2-8} 
\multicolumn{1}{c}{} & Text Expansion & \multicolumn{1}{c}{8.18} & \multicolumn{1}{c}{8.12} & 99.28 & \multicolumn{1}{c}{8.47} & \multicolumn{1}{c}{8.41} & 99.31 \\ \cmidrule{2-8} 
\multicolumn{1}{c}{} & Text Creation & \multicolumn{1}{c}{6.82} & \multicolumn{1}{c}{6.55} & 96.16 & \multicolumn{1}{c}{6.50} & \multicolumn{1}{c}{6.42} & 98.82 \\ \cmidrule{2-8} 
\multicolumn{1}{c}{} & Text Formatting & \multicolumn{1}{c}{6.10} & \multicolumn{1}{c}{5.67} & 92.97 & \multicolumn{1}{c}{4.33} & \multicolumn{1}{c}{3.99} & 91.97 \\ \cmidrule{2-8} 
\multicolumn{1}{c}{} & Instant Reply & \multicolumn{1}{c}{6.55} & \multicolumn{1}{c}{6.30} & 96.18 & \multicolumn{1}{c}{6.20} & \multicolumn{1}{c}{5.94} & 95.84 \\ \cmidrule{2-8} 
\multicolumn{1}{c}{} & Text Correction & \multicolumn{1}{c}{2.83} & \multicolumn{1}{c}{2.24} & 78.89 & \multicolumn{1}{c}{1.67} & \multicolumn{1}{c}{1.17} & 70.00 \\ \midrule
\multicolumn{1}{c}{\multirow{3}{*}{Information   Extraction}} & Entity Extraction & \multicolumn{1}{c}{7.15} & \multicolumn{1}{c}{7.05} & 98.49 & \multicolumn{1}{c}{6.13} & \multicolumn{1}{c}{6.08} & 99.06 \\ \cmidrule{2-8} 
\multicolumn{1}{c}{} & Relation Extraction & \multicolumn{1}{c}{5.73} & \multicolumn{1}{c}{5.01} & 87.48 & \multicolumn{1}{c}{4.64} & \multicolumn{1}{c}{3.62} & 78.06 \\ \cmidrule{2-8} 
\multicolumn{1}{c}{} & Event Extraction & \multicolumn{1}{c}{7.00} & \multicolumn{1}{c}{6.06} & 86.68 & \multicolumn{1}{c}{7.06} & \multicolumn{1}{c}{6.05} & 85.62 \\ \midrule
\multicolumn{1}{c}{\multirow{2}{*}{Notification   Management}} & Message Summ & \multicolumn{1}{c}{7.92} & \multicolumn{1}{c}{7.80} & 98.48 & \multicolumn{1}{c}{8.00} & \multicolumn{1}{c}{7.88} & 98.50 \\ \cmidrule{2-8} 
\multicolumn{1}{c}{} & Notification Sorting & \multicolumn{1}{c}{5.13} & \multicolumn{1}{c}{4.83} & 94.14 & \multicolumn{1}{c}{4.90} & \multicolumn{1}{c}{4.74} & 96.71 \\ \midrule
\rowcolor{gray!20}
\multicolumn{2}{c}{AVG} & \multicolumn{1}{c}{6.96} & \multicolumn{1}{c}{6.35} & 91.31 & \multicolumn{1}{c}{6.71} & \multicolumn{1}{c}{6.08} & 90.58 \\ \midrule
\end{tabular}%
}
\vspace{-0.5em}
\caption{W4A16 evaluation results with 50 questions per task using GPT-4 Turbo as the judge LLM. We deploy BlueLM-3B and Qwen2.5-3B on the NPU of the vivo iQOO 12 smartphone, which is equipped with the Snapdragon 8 Gen 3 SoC. The quantized models are able to maintain an average performance of around 90\%.}
\label{tab:int4gpt}
\vspace{-1em}
\end{table*}

\subsection{Evaluation Protocol}\label{sec:eval_pro}

Since subjective questions often lack an absolutely correct answer and involve multifaceted scoring dimensions, current subjective question evaluation datasets always adopt the ``LLM-as-a-Judge'' approach for assessment~\cite{liu2023alignbench,zheng2023judging,arenahard2024}. In SmartBench, we meticulously design different LLM evaluation prompts for each function category. For Content Creation, Information Extraction, and Notification Management, we especially design distinct scoring prompts for each task. This targeted design makes the scoring more aligned with human perceptions. 

In SmartBench, each question is assigned a total score of 10 points. For the evaluation prompt of each task, in addition to providing reference answers for each question, we also include detailed scoring guidelines. We first outline the scoring dimensions; for example, in the text continuation task (as in Fig.~\ref{fig:score_prompt}), we assess the answer's coherence, language quality, creativity, and consistency with the original text. Next, we develop comprehensive scoring standards for each dimension to ensure accurate and consistent grading. The judge LLM first assigns separate scores for each dimension and then provides an overall aggregate score. Especially, for the Text Correction task, which has clearly defined correction answers, the evaluation criterion focuses on the accuracy of the modifications made.

\section{Experiment}\label{sec:exp}
 
In this section, a series of experiments are conducted. We evaluate the performance of representative on-device LLMs and MLLMs on SmartBench (Sec.~\ref{sec:full_fp}) and conduct human tests to assess the effectiveness of the LLM-as-a-Judge evaluation method (Sec.~\ref{sec:human_test}). To better align with practical on-device deployment, we also analyze the model performance after quantized inference on the NPU in actual smartphones (Sec.~\ref{sec:int4_res}).

\subsection{BF16 Precision Evaluation}\label{sec:full_fp}

In this subsection, we evaluate representative on-device LLMs/MLLMs on SmartBench (BF16 parameter precision). We select BlueLM-3B~\cite{lu2024bluelm}, InternVL2.5-4B~\cite{chen2024expanding}, MiniCPM3-4B~\cite{hu2024minicpm}, Qwen2.5-3B~\cite{yang2024qwen2}, and Qwen2-VL-2B~\cite{Qwen2VL}. GPT-4 Turbo (\texttt{gpt-4-turbo-04-09}) is utilized as the judge LLM. We also provide the scores of GPT-4o to compare on-cloud models with on-device models. The results are summarized in Tab.~\ref{tab:fp}, where BlueLM-3B achieves the highest average score among on-device models. Additionally, we can observe the following trends from the table:

\textbf{1)} For common text-based tasks on mobile devices, such as summarization and question-answering, existing on-device models have shown satisfactory performance. However, when dealing with tasks that require more rigorous logical reasoning, such as Text Correction, Relation Extraction, and Notification Sorting, the performance of on-device models still lags behind. We provide several examples in the Appendix. For instance, Fig.~\ref{fig:example2} demonstrates that \underline{all models struggle} to identify subtle typos within sentences.

\begin{table*}[t]
\vspace{-3em}
\renewcommand{\arraystretch}{0.1}
\fontsize{8pt}{10pt}\selectfont
\begin{tabular*}{\textwidth}{@{\extracolsep{\fill}}lccccc}
\midrule
 & \textbf{\#Params} & \textbf{Context Length} & \textbf{Prefilling Speed (token/s)} & \textbf{Output Speed (token/s)} & \textbf{Power (W)} \\
\midrule
{BlueLM-3B} & 2.7B & 2048 & 930.9 & 27.1 & 6.4 \\ \midrule
{Qwen2.5-3B} & 3.1B & 2048 & 873.4 & 24.9 & 6.8 \\
\midrule
\end{tabular*}
\vspace{-1.2em}
\caption{Inference speed and power usage of BlueLM-3B and Qwen2.5-3B on iQOO 12 with Qualcomm QNN SDK. Due to its larger parameter size, Qwen2.5-3B exhibits slower inference speed and higher power consumption.}
\label{tab:speed}
\end{table*}

\begin{table*}[t]\small
\vspace{-1em}
\fontsize{11pt}{10pt}\selectfont
\renewcommand\arraystretch{0.6}
\resizebox{\textwidth}{!}{%
\begin{tabular}{ccccccc}
\midrule
 & \textbf{Text Summarization} & \textbf{Text Q\&A} & \textbf{Content Creation} & \textbf{Information Extraction} & \textbf{Notification Management} & \textbf{AVG} \\ \midrule
MT-Bench & 0.8412 & 0.8025 & 0.6998 & 0.7894 & 0.8467 & 0.7959 \\ \midrule
\rowcolor{gray!20}
SmartBench & 0.8823 & 0.8151 & 0.7289 & 0.8396 & 0.8742 & 0.8280 \\ \midrule
\end{tabular}%
}
\vspace{-0.5em}
\caption{We compare our LLM-as-a-Judge evaluation method with MT-Bench's evaluation method using the Pearson correlation score with human rankings. Our evaluation method demonstrates higher consistency with humans.}
\label{tab:human_test}
\vspace{-1em}
\end{table*}

\textbf{2)} Integrating multimodal capabilities into MLLMs might result in a reduction of pure language performance. Specifically, the InternVL2.5-4B model is developed based on Qwen2.5-3B. While InternVL2.5-4B successfully acquires multimodal functionalities, this enhancement leads to a partial decline in its pure language performance.

\textbf{3)} On-device models still exhibit notable performance gaps compared to the on-cloud model, particularly on the Text Correction task, where they achieve only about half the score of GPT-4o. This suggests that enhancing the reasoning capability of on-device models remains an important area.

\textbf{4)} For a more comprehensive evaluation, we also present the scores assessed by Qwen-Max (\texttt{qwen-max-longcontext}) as the judging LLM in Tab.~\ref{tab:fp_qwen} in the Appendix. It can be observed that although there are slight differences in the average scores, both Qwen-Max and GPT-4 Turbo rank the models in the same order. This demonstrates the robustness of our LLM-as-a-Judge approach.

\textbf{\underline{Remark:}} We evaluate MLLMs on SmartBench because in on-device deployment scenarios on real smartphones, memory limitations often prevent us from deploying both an LLM and an MLLM on the device. Consequently, this on-device model must simultaneously handle both pure language tasks and multimodal tasks effectively.

\subsection{INT4 Precision Evaluation on NPU}\label{sec:int4_res}

On-device LLMs are often deployed on the smartphone's Neural Processing Unit (NPU) to leverage its specialized parallel computational capabilities. In our experiment, we deploy the BlueLM-3B and Qwen2.5-3B models on the NPU of the vivo iQOO 12 smartphone equipped with the Snapdragon 8 Gen 3 SoC. To be specific, we quantize the models to W4A16 using the Qualcomm QNN SDK. Due to the inference speed limitations on the mobile NPU, we select 50 questions per task for inference. The results are shown in Tab.~\ref{tab:int4gpt}. We present the scores for each task (BF16 and INT4) and the capability retention of the INT4 models. Additionally, we provide the evaluation results using Qwen-Max as the judge LLM in Tab.~\ref{tab:int4qw} in the Appendix.

\textbf{1)} For most tasks, the quantized models retain over 80\% of their original capabilities, with an overall average retention rate of approximately 90\%. 

\textbf{2)} Although the models can achieve, on average, 90\% of the original score on the edge side, they may still generate incorrect responses after quantization. We provide two failure cases of BlueLM-3B in Sec.~\ref{sec:fail_quant}, where the model exhibits fluency degradation and reduced understanding capability.

\textbf{3)} To offer deeper insights into real-world edge-side deployment, we report the prefilling speed, output token generation speed, and power consumption on the iQOO 12 smartphone using the Qualcomm QNN SDK, as shown in Tab.~\ref{tab:speed}.

\subsection{Human Test}\label{sec:human_test}

We use the LLM-as-a-Judge method to assess different on-device models. Therefore, it is important to examine the consistency between the scores given by the judge LLM and those given by humans. We carry out a human test with six human experts in this subsection.

During the auto-evaluation process, the judge LLM assigns a score between 0 and 10 to the output of each model response. Considering that humans might find it challenging to directly score subjective questions, especially tasks like text polishing, we ask human experts to rank the outputs generated by different on-device models (i.e., BlueLM-3B, InternVL2.5-4B, MiniCPM3-4B, Qwen2.5-3B, and Qwen2-VL-2B) for each question. We then use the scores from the judge LLM (Qwen-Max in our setting) to compute model rankings for each question. Finally, we calculate the Pearson correlation between the rankings from the judge LLM and those provided by human experts.

In SmartBench, we meticulously design evaluation dimensions and scoring standards for each task/category. To establish a baseline, we compare our evaluation prompts with those used in MT-Bench. We randomly select 50 questions for each task, with each question containing responses from 5 on-device models. This results in a total of 50$\times$20$\times$5=5000 samples. We conduct human ranking and calculate the Pearson correlation with the judge LLM ranking (our prompt versus MT-Bench prompt), and the results are shown in Tab.~\ref{tab:human_test}. Our designed prompt excels in all categories.

\section{Conclusion}\label{sec:conclusion}

In this paper, we present SmartBench, the first benchmark designed to evaluate the capabilities of on-device LLMs in Chinese mobile contexts. By analyzing functionalities offered by leading smartphone manufacturers, we create a standardized framework divided into five key categories and 20 specific tasks, complete with high-quality datasets and tailored evaluation criteria. Our comprehensive evaluations of on-device LLMs and MLLMs using SmartBench highlight the strengths and weaknesses of current models in real-world mobile scenarios. This work fills a critical gap in benchmarking tools for Chinese users, promoting further development and optimization of on-device LLMs in practical mobile applications.

\section*{Limitations}

In this paper, we provide SmartBench, the first benchmark designed to evaluate the capabilities of on-device LLMs in Chinese mobile contexts. Our work still has some limitations: 1) With the advancement of technology, the functions of on-device LLMs will continually evolve. Our investigation only covers up to December 2024. We will continue to update the dataset in line with the release of new features. 2) We have developed SmartBench specifically for the usage scenarios of Chinese users. The usage habits and methods of smartphone users may vary significantly across different countries. Moving forward, we will continue to support multiple languages. 3) The current benchmark focuses on the text modality, whereas mobile applications may also involve vision and audio modalities (e.g., camera input and voice recognition). We plan to incorporate these additional modalities in future versions.

\bibliography{custom}
\clearpage
\appendix
\section{Appendix}
\label{sec:appendix}

\subsection{Data License}

\begin{table}[h]
\vspace{-0.5em}
\resizebox{\textwidth}{!}{%
\renewcommand\arraystretch{0.6}
\begin{tabular}{lll}
\midrule
\textbf{Dataset} & \textbf{Source} & \textbf{License} \\ \midrule
nlp\_chinese\_corpus & https://github.com/brightmart/nlp\_chinese\_corpus & MIT License \\ \midrule
WenetSpeech & https://wenet.org.cn/WenetSpeech/ & CC BY 4.0 \\ \midrule
LCCC & https://github.com/thu-coai/CDial-GPT & MIT License \\ \midrule
Alimeeting4MUG & https://modelscope.cn/datasets/modelscope/Alimeeting4MUG/ & CC BY 4.0 \\ \midrule
VCSum & https://github.com/hahahawu/VCSum & MIT License \\ \midrule
CMRC 2018 & https://ymcui.com/cmrc2018/ & CC BY-SA 4.0 \\ \midrule
DuReader-2.0 & https://github.com/baidu/DuReader/tree/master/DuReader-2.0 & Apache License 2.0 \\ \midrule
Weibo & https://github.com/hltcoe/golden-horse & CC BY-SA 3.0 \\ \midrule
MSRA & https://tianchi.aliyun.com/dataset/144307 & CC BY 4.0 \\ \midrule
OntoNotes Release 4.0 & https://www.modelscope.cn/datasets/yingxi/cross\_ner & Apache License 2.0 \\ \midrule
CSCD-NS & https://github.com/nghuyong/cscd-ns & MIT License \\ \midrule
\end{tabular}%
}
\vspace{-0.5em}
\caption{\parbox{\textwidth}{Data license of the open-source datasets used in SmartBench.}}
\label{tab:license}
\vspace{-1.5em}
\end{table}

\subsection{More Evaluation Results}

\begin{table}[h]
\vspace{-1em}
\resizebox{\textwidth}{!}{%
\begin{tabular}{clccccc}
\midrule
\multicolumn{1}{c}{\textbf{}} & \textbf{Task} & \textbf{BlueLM-3B} & \textbf{InternVL2.5-4B} & \textbf{MiniCPM3-4B} & \textbf{Qwen2.5-3B} & \textbf{Qwen2-VL-2B} \\ \midrule
\multicolumn{1}{c}{\multirow{4}{*}{\begin{tabular}[c]{@{}c@{}}Text\\      Summarization\end{tabular}}} & Document Summ & 7.20 & 6.98 & 6.95 & 7.09 & 4.74 \\ \cmidrule{2-7} 
\multicolumn{1}{c}{} & Call Summ & 7.27 & 5.97 & 6.94 & 6.58 & 4.12 \\ \cmidrule{2-7} 
\multicolumn{1}{c}{} & Recording Summ & 7.13 & 6.56 & 7.02 & 7.00 & 4.58 \\ \cmidrule{2-7} 
\multicolumn{1}{c}{} & Meeting Summ & 7.22 & 6.70 & 7.10 & 7.07 & 4.36 \\ \midrule
\multicolumn{1}{c}{\multirow{3}{*}{Text Q\&A}} & Document Q\&A & 8.45 & 8.46 & 7.79 & 8.63 & 8.34 \\ \cmidrule{2-7} 
\multicolumn{1}{c}{} & Retrieval Q\&A & 6.14 & 5.95 & 6.83 & 6.33 & 4.92 \\ \cmidrule{2-7} 
\multicolumn{1}{c}{} & Personal Q\&A & 8.37 & 8.30 & 8.04 & 8.57 & 8.16 \\ \midrule
\multicolumn{1}{c}{\multirow{8}{*}{\begin{tabular}[c]{@{}c@{}}Content\\      Creation\end{tabular}}} & Text Polishing & 6.93 & 5.78 & 6.90 & 6.86 & 5.91 \\ \cmidrule{2-7} 
\multicolumn{1}{c}{} & Text Continuation & 7.13 & 6.56 & 7.19 & 7.31 & 5.60 \\ \cmidrule{2-7} 
\multicolumn{1}{c}{} & Text Abbreviation & 7.23 & 6.72 & 7.40 & 7.63 & 6.52 \\ \cmidrule{2-7} 
\multicolumn{1}{c}{} & Text Expansion & 6.79 & 7.04 & 7.23 & 7.31 & 5.72 \\ \cmidrule{2-7} 
\multicolumn{1}{c}{} & Text Creation & 6.73 & 5.78 & 6.67 & 6.63 & 5.02 \\ \cmidrule{2-7} 
\multicolumn{1}{c}{} & Text Formatting & 6.35 & 5.93 & 7.03 & 5.25 & 2.93 \\ \cmidrule{2-7} 
\multicolumn{1}{c}{} & Instant Reply & 5.60 & 5.09 & 5.25 & 5.26 & 3.17 \\ \cmidrule{2-7} 
\multicolumn{1}{c}{} & Text Correction & 3.53 & 2.39 & 3.48 & 2.06 & 1.24 \\ \midrule
\multicolumn{1}{c}{\multirow{3}{*}{\begin{tabular}[c]{@{}c@{}}Information\\      Extraction\end{tabular}}} & Entity Extraction & 7.77 & 7.40 & 7.44 & 6.21 & 5.00 \\ \cmidrule{2-7} 
\multicolumn{1}{c}{} & Relation Extraction & 5.73 & 4.13 & 4.77 & 3.97 & 3.65 \\ \cmidrule{2-7} 
\multicolumn{1}{c}{} & Event Extraction & 7.14 & 7.32 & 6.85 & 7.21 & 5.17 \\ \midrule
\multicolumn{1}{c}{\multirow{2}{*}{\begin{tabular}[c]{@{}c@{}}Notification\\      Management\end{tabular}}} & Message Summ & 6.92 & 6.96 & 7.47 & 7.62 & 5.64 \\ \cmidrule{2-7} 
\multicolumn{1}{c}{} & Notification Sorting & 5.38 & 4.63 & 5.56 & 5.21 & 2.93 \\ \midrule
\rowcolor{gray!20}
\multicolumn{2}{c}{AVG} & 6.75 & 6.23 & 6.70 & 6.49 & 4.89 \\ \midrule
\end{tabular}%
}
\caption{\parbox{\textwidth}{Evaluation results using Qwen-Max (\texttt{qwen-max-longcontext}) as the judge LLM with BF16 precision.}}
\label{tab:fp_qwen}
\vspace{-0.5em}
\end{table}

We present the scores evaluated by Qwen-Max as the judging LLM in Tab.~\ref{tab:fp_qwen} with BF16 precision. When compared to the GPT-4 Turbo results shown in Tab.~\ref{tab:fp}, both Qwen-Max and GPT-4 Turbo rank the models in the same order. This demonstrates the robustness of our LLM-as-a-Judge approach. 

\newpage
\vspace*{56em}

We also include the INT4 precision inference performance (evaluated by Qwen-Max) of BlueLM-3B and Qwen2.5-3B on the vivo iQOO 12 smartphone (50 questions per task), along with the performance retention compared to the original BF16 models. As shown in Tab.~\ref{tab:int4qw}.

\clearpage
\onecolumn
\begin{table}[h]
\vspace{-1em}
\renewcommand\arraystretch{0.75}
\resizebox{\textwidth}{!}{%
\resizebox{\textwidth}{!}{%
\begin{tabular}{clcccccc}
\midrule
\multicolumn{1}{c}{\textbf{Category}} & \multicolumn{1}{c}{\textbf{Task}} & \multicolumn{3}{c}{\textbf{BlueLM-3B}} & \multicolumn{3}{c}{\textbf{Qwen2.5-3B}} \\ \midrule
\multicolumn{2}{c}{\textbf{Precision}} & \multicolumn{1}{c}{\textbf{BF16}} & \multicolumn{1}{c}{\textbf{INT4}} & \textbf{Retention (\%)} & \multicolumn{1}{c}{\textbf{BF16}} & \multicolumn{1}{c}{\textbf{INT4}} & \textbf{Retention (\%)} \\ \midrule
\multicolumn{1}{c}{\multirow{4}{*}{Text   Summarization}} & Document Summ & \multicolumn{1}{c}{6.89} & \multicolumn{1}{c}{5.62} & 81.61 & \multicolumn{1}{c}{6.67} & \multicolumn{1}{c}{4.00} & 60.00 \\ \cmidrule{2-8} 
\multicolumn{1}{c}{} & Call Summ & \multicolumn{1}{c}{6.71} & \multicolumn{1}{c}{5.89} & 87.66 & \multicolumn{1}{c}{6.57} & \multicolumn{1}{c}{5.43} & 82.61 \\ \cmidrule{2-8} 
\multicolumn{1}{c}{} & Recording Summ & \multicolumn{1}{c}{6.95} & \multicolumn{1}{c}{6.63} & 95.38 & \multicolumn{1}{c}{6.95} & \multicolumn{1}{c}{6.53} & 94.03 \\ \cmidrule{2-8} 
\multicolumn{1}{c}{} & Meeting Summ & \multicolumn{1}{c}{7.39} & \multicolumn{1}{c}{5.57} & 75.29 & \multicolumn{1}{c}{7.10} & \multicolumn{1}{c}{5.10} & 71.84 \\ \midrule
\multicolumn{1}{c}{\multirow{3}{*}{Text Q\&A}} & Document Q\&A & \multicolumn{1}{c}{8.50} & \multicolumn{1}{c}{8.32} & 97.83 & \multicolumn{1}{c}{8.85} & \multicolumn{1}{c}{8.77} & 99.13 \\ \cmidrule{2-8} 
\multicolumn{1}{c}{} & Retrieval Q\&A & \multicolumn{1}{c}{6.19} & \multicolumn{1}{c}{5.75} & 92.93 & \multicolumn{1}{c}{6.31} & \multicolumn{1}{c}{6.13} & 97.03 \\ \cmidrule{2-8} 
\multicolumn{1}{c}{} & Personal Q\&A & \multicolumn{1}{c}{8.06} & \multicolumn{1}{c}{7.16} & 88.80 & \multicolumn{1}{c}{8.42} & \multicolumn{1}{c}{8.26} & 98.08 \\ \midrule
\multicolumn{1}{c}{\multirow{8}{*}{Content Creation}} & Text Polishing & \multicolumn{1}{c}{6.89} & \multicolumn{1}{c}{6.75} & 97.93 & \multicolumn{1}{c}{6.82} & \multicolumn{1}{c}{6.39} & 93.72 \\ \cmidrule{2-8} 
\multicolumn{1}{c}{} & Text Continuation & \multicolumn{1}{c}{7.17} & \multicolumn{1}{c}{7.07} & 98.60 & \multicolumn{1}{c}{7.17} & \multicolumn{1}{c}{7.03} & 98.14 \\ \cmidrule{2-8} 
\multicolumn{1}{c}{} & Text Abbreviation & \multicolumn{1}{c}{7.26} & \multicolumn{1}{c}{7.10} & 97.78 & \multicolumn{1}{c}{7.61} & \multicolumn{1}{c}{7.58} & 99.58 \\ \cmidrule{2-8} 
\multicolumn{1}{c}{} & Text Expansion & \multicolumn{1}{c}{7.24} & \multicolumn{1}{c}{7.13} & 98.54 & \multicolumn{1}{c}{7.65} & \multicolumn{1}{c}{7.29} & 95.38 \\ \cmidrule{2-8} 
\multicolumn{1}{c}{} & Text Creation & \multicolumn{1}{c}{6.96} & \multicolumn{1}{c}{6.48} & 93.15 & \multicolumn{1}{c}{6.62} & \multicolumn{1}{c}{6.23} & 94.19 \\ \cmidrule{2-8} 
\multicolumn{1}{c}{} & Text Formatting & \multicolumn{1}{c}{6.45} & \multicolumn{1}{c}{6.38} & 98.93 & \multicolumn{1}{c}{5.34} & \multicolumn{1}{c}{5.00} & 93.55 \\ \cmidrule{2-8} 
\multicolumn{1}{c}{} & Instant Reply & \multicolumn{1}{c}{5.60} & \multicolumn{1}{c}{5.21} & 93.04 & \multicolumn{1}{c}{4.95} & \multicolumn{1}{c}{4.15} & 83.84 \\ \cmidrule{2-8} 
\multicolumn{1}{c}{} & Text Correction & \multicolumn{1}{c}{3.17} & \multicolumn{1}{c}{2.00} & 63.16 & \multicolumn{1}{c}{1.83} & \multicolumn{1}{c}{1.17} & 63.64 \\ \midrule
\multicolumn{1}{c}{\multirow{3}{*}{Information   Extraction}} & Entity Extraction & \multicolumn{1}{c}{6.98} & \multicolumn{1}{c}{6.66} & 95.46 & \multicolumn{1}{c}{5.65} & \multicolumn{1}{c}{5.34} & 94.40 \\ \cmidrule{2-8} 
\multicolumn{1}{c}{} & Relation Extraction & \multicolumn{1}{c}{5.64} & \multicolumn{1}{c}{4.77} & 84.63 & \multicolumn{1}{c}{4.12} & \multicolumn{1}{c}{3.81} & 92.50 \\ \cmidrule{2-8} 
\multicolumn{1}{c}{} & Event Extraction & \multicolumn{1}{c}{6.90} & \multicolumn{1}{c}{6.10} & 88.32 & \multicolumn{1}{c}{7.32} & \multicolumn{1}{c}{6.80} & 92.83 \\ \midrule
\multicolumn{1}{c}{\multirow{2}{*}{Notification   Management}} & Message Summ & \multicolumn{1}{c}{7.16} & \multicolumn{1}{c}{7.12} & 99.44 & \multicolumn{1}{c}{7.60} & \multicolumn{1}{c}{7.52} & 98.95 \\ \cmidrule{2-8} 
\multicolumn{1}{c}{} & Notification Sorting & \multicolumn{1}{c}{5.50} & \multicolumn{1}{c}{5.34} & 96.95 & \multicolumn{1}{c}{5.35} & \multicolumn{1}{c}{5.15} & 96.20 \\ \midrule
\rowcolor{gray!20}
\multicolumn{2}{c}{AVG} & \multicolumn{1}{c}{6.68} & \multicolumn{1}{c}{6.15} & 92.08 & \multicolumn{1}{c}{6.45} & \multicolumn{1}{c}{5.88} & 91.29 \\ \midrule
\end{tabular}%
}
}
\caption{\parbox{\textwidth}{Evaluation results using Qwen-Max (\texttt{qwen-max-longcontext}) as the judge LLM with INT4 precision.}}
\label{tab:int4qw}
\end{table}

\subsection{Details of Human Annotators}

In the Data Screening and Human Test stages, we hire six domain experts with over five years of mobile AI experience. These experts have at least a master's degree. We pay them a labeling fee of \$20 per hour.

\subsection{Comparison with Traditional Benchmarks}

Traditional benchmarks (e.g., MMLU) mainly evaluate the model’s objective knowledge, while SmartBench focuses on subjective data aligned with end-side smartphone scenarios, assessing the degree of alignment with human preferences. In practice, due to memory and storage constraints on mobile devices, only a single MLLM can be deployed on the device. We consider the training process from LLM to MLLM. For example, InternVL2.5-4B is trained from Qwen2.5-3B-Instruct. During this process, objective knowledge tends to be preserved, but the subjective (human alignment) performance often degrades.

\begin{table}[h]\small
\centering
\begin{tabular}{lccc}
\midrule
\textbf{} & \textbf{MMLU} & \textbf{SmartBench (GPT-4)} & \textbf{SmartBench (Qwen-Max)} \\ \midrule
\textbf{Qwen2.5-3B} & 66.31 & 6.61 & 6.49 \\ \midrule
\textbf{InternVL2.5-4B} & 68.35 & 6.37 & 6.23 \\ \midrule
\end{tabular}
\end{table}

MMLU measures objective knowledge, while SmartBench evaluates subjective alignment. We test SmartBench performance using both GPT-4 and Qwen-Max as evaluators. As can be seen, during the training process from LLM to multimodal models, MMLU performance improves due to the injection of new knowledge. However, the subjective performance represented by SmartBench decreases.

\onecolumn
\clearpage

\subsection{Failure Cases on NPU}\label{sec:fail_quant}

We here provide two failure cases of BlueLM-3B after quantized deployment on the NPU.

\begin{figure*}[ht]
    \centering
    \includegraphics[width=0.9\linewidth]{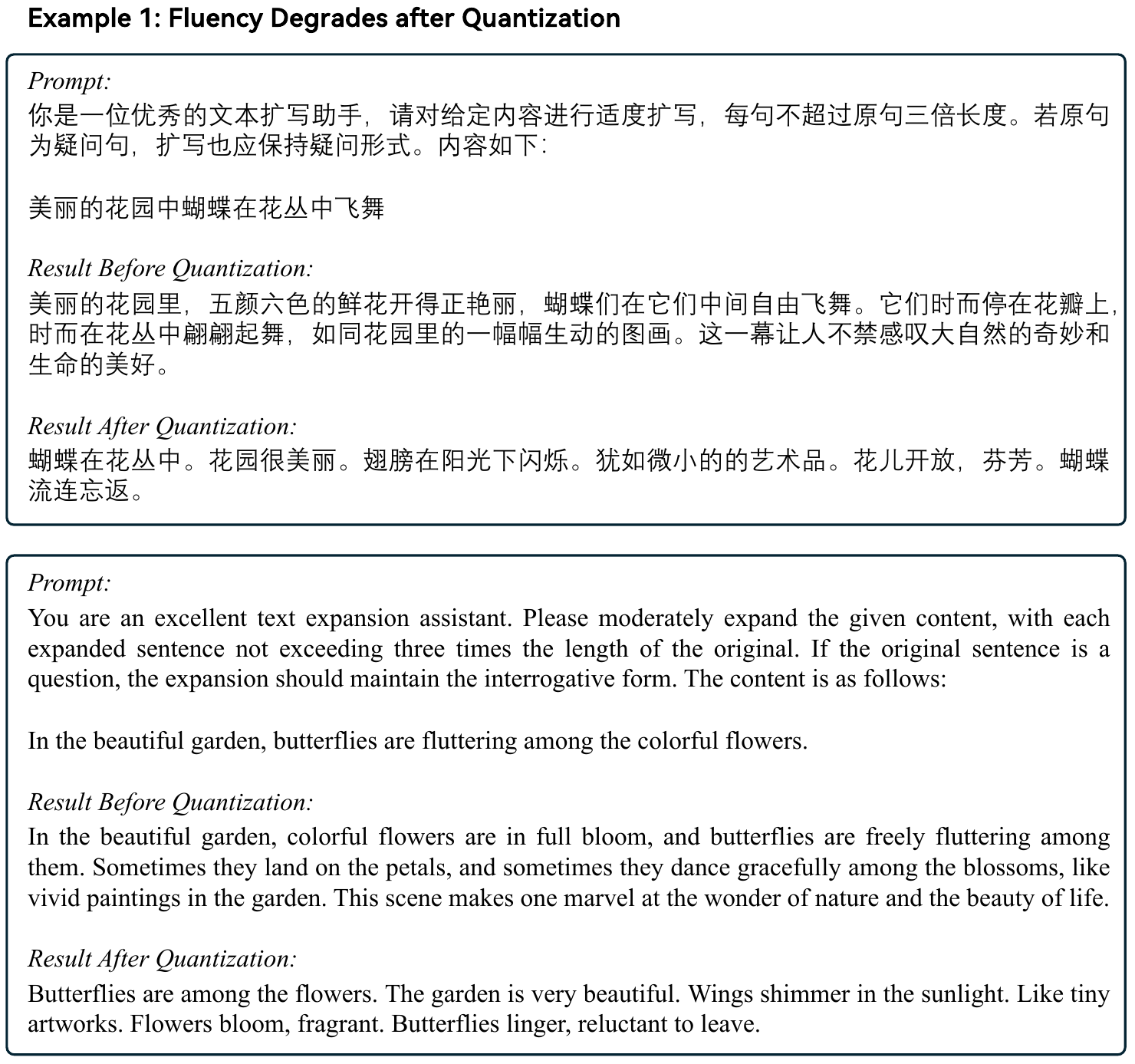}
    \caption{Failure case after quantization on the NPU.}
    \label{fig:example5}
\end{figure*}

\begin{figure*}[ht]
    \centering
    \includegraphics[width=0.9\linewidth]{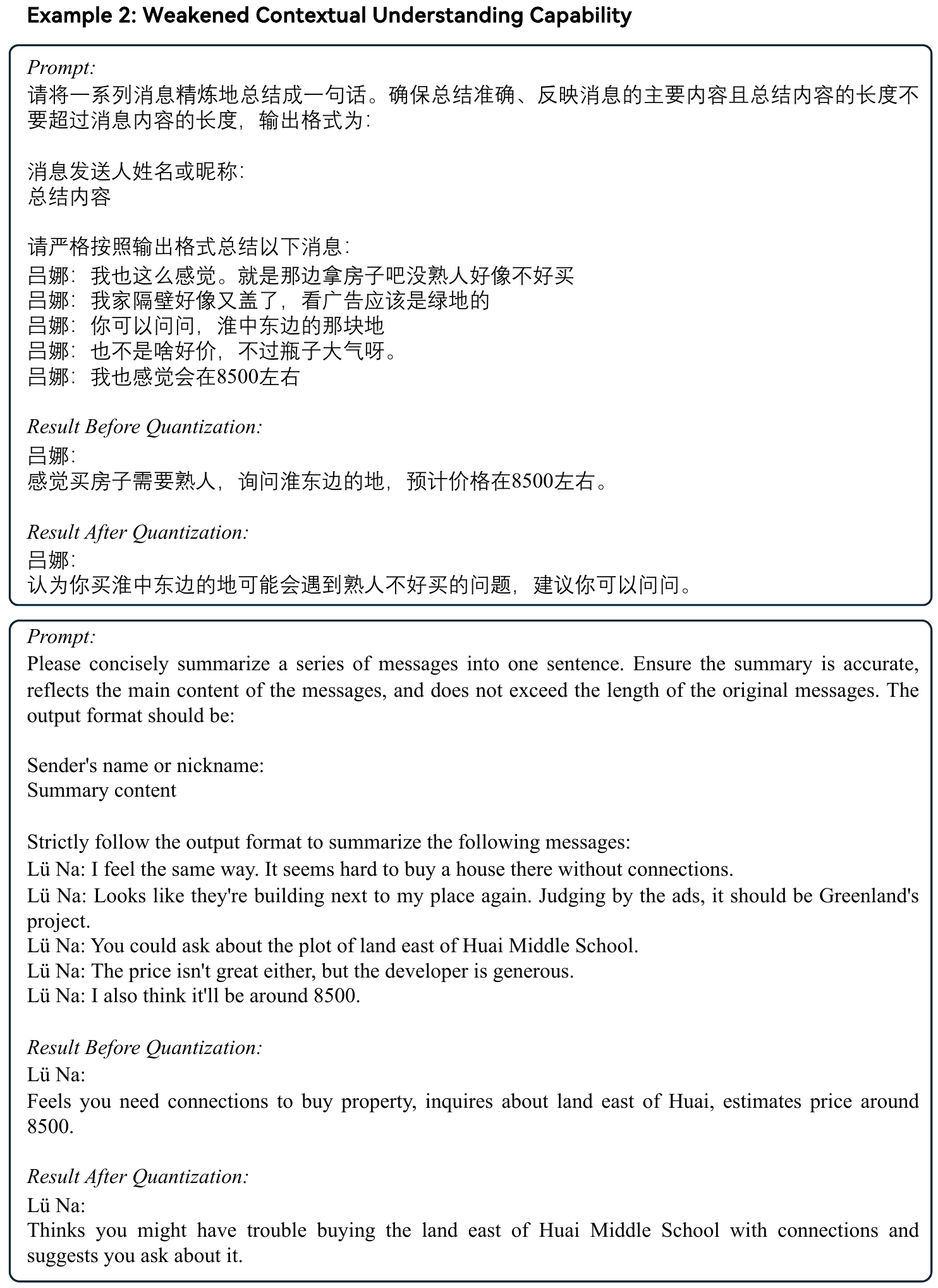}
    \caption{Failure case after quantization on the NPU.}
    \label{fig:example6}
\end{figure*}

\clearpage
\subsection{Example Prompt for Generating Task Queries}

\begin{figure}[h]
    \centering
    \includegraphics[width=\linewidth]{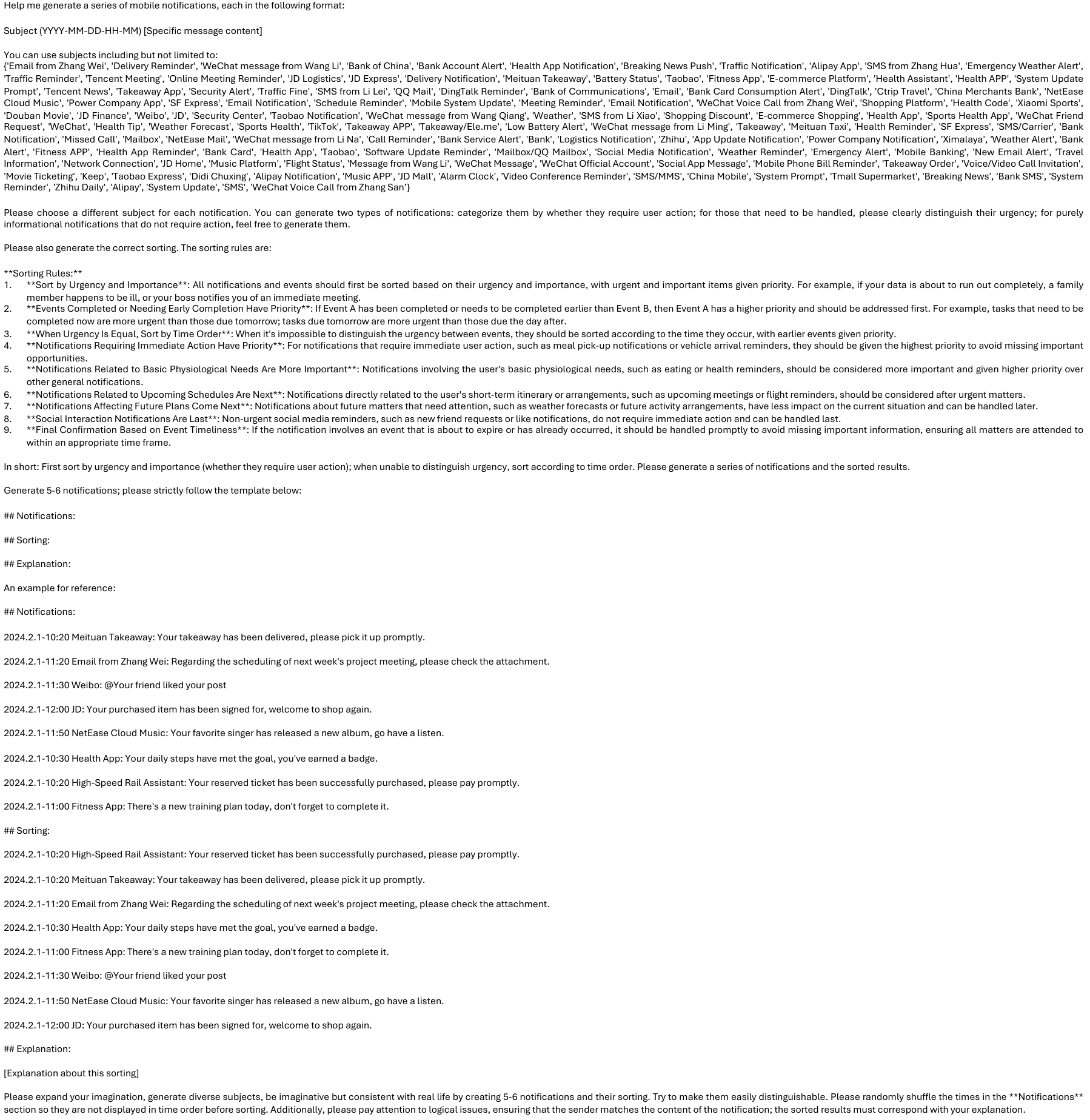}
    \vspace{-1em}
    \caption{Prompt for generating mobile notification sorting pairs (en).}
    \label{fig:prompt_sort}
\end{figure}

\clearpage

\subsection{More Example Cases}

We provide examples of SmartBench, along with the inference results of different models using BF16 precision (zh). The corresponding English translation is also provided (en). The numbers in red represent the scores given by Qwen-Max (\texttt{qwen-max-longcontext}).

\begin{figure*}[ht]
    \centering
    \includegraphics[width=0.85\linewidth]{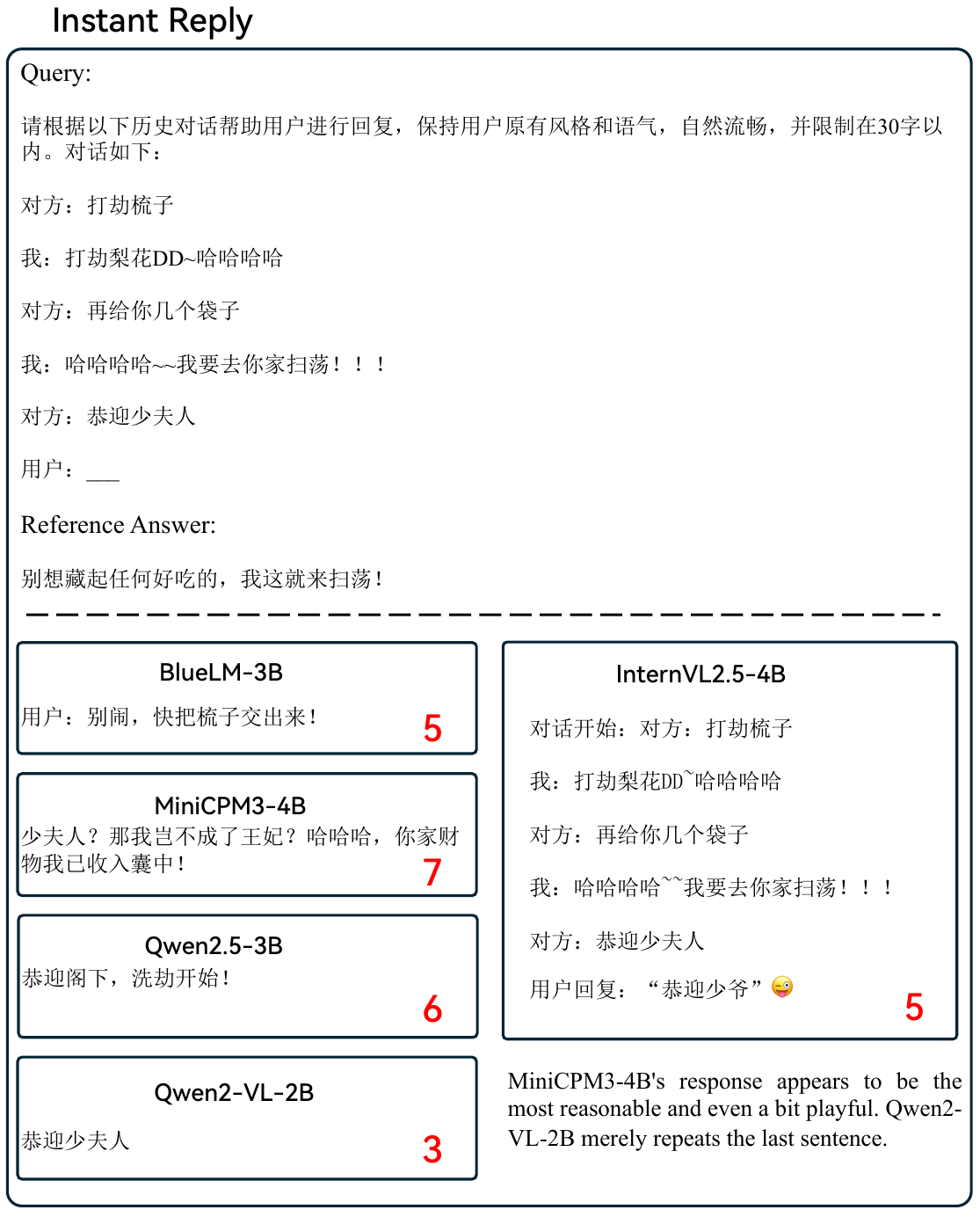}
    \caption{Instant reply case (zh).}
    \label{fig:example1}
\end{figure*}

\begin{figure*}[ht]
    \centering
    \includegraphics[width=0.85\linewidth]{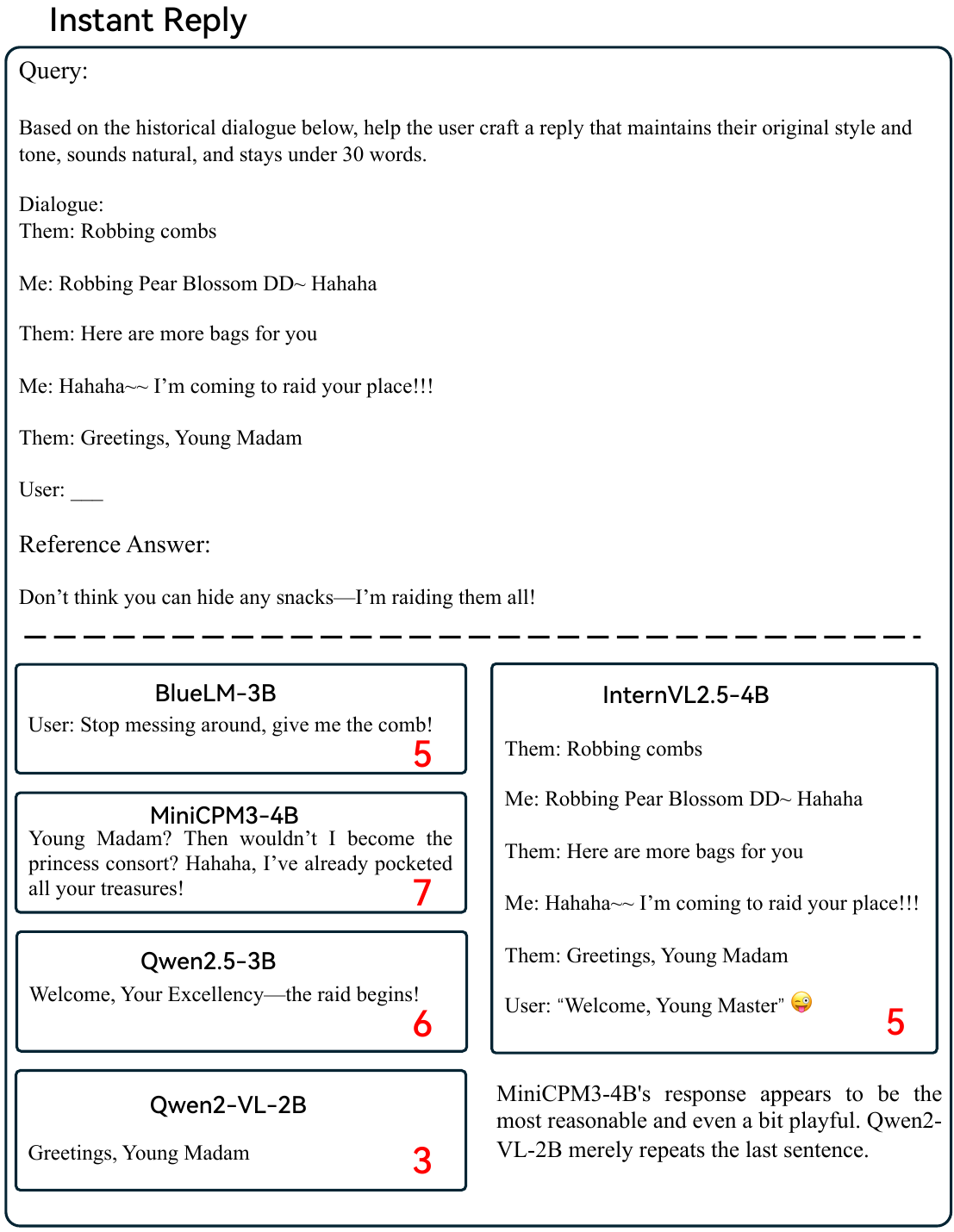}
    \caption{Instant reply case (en).}
    \label{fig:example1_en}
\end{figure*}

\begin{figure}[htbp]
    \centering
    \includegraphics[width=0.85\linewidth]{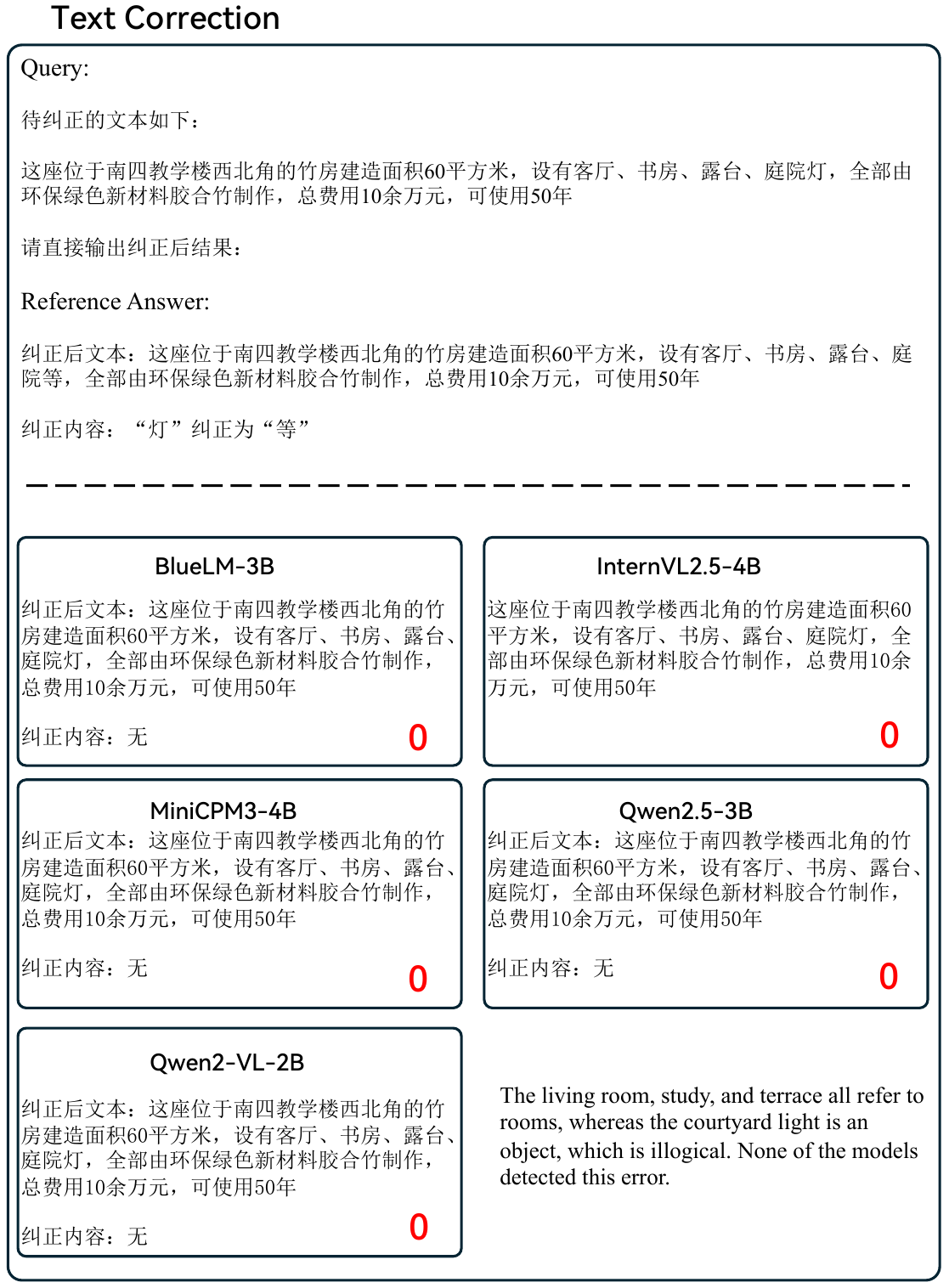}
    \caption{Text correction case (zh).}
    \label{fig:example2}
\end{figure}

\begin{figure}[htbp]
\vspace{-1em}
    \centering
    \includegraphics[width=0.85\linewidth]{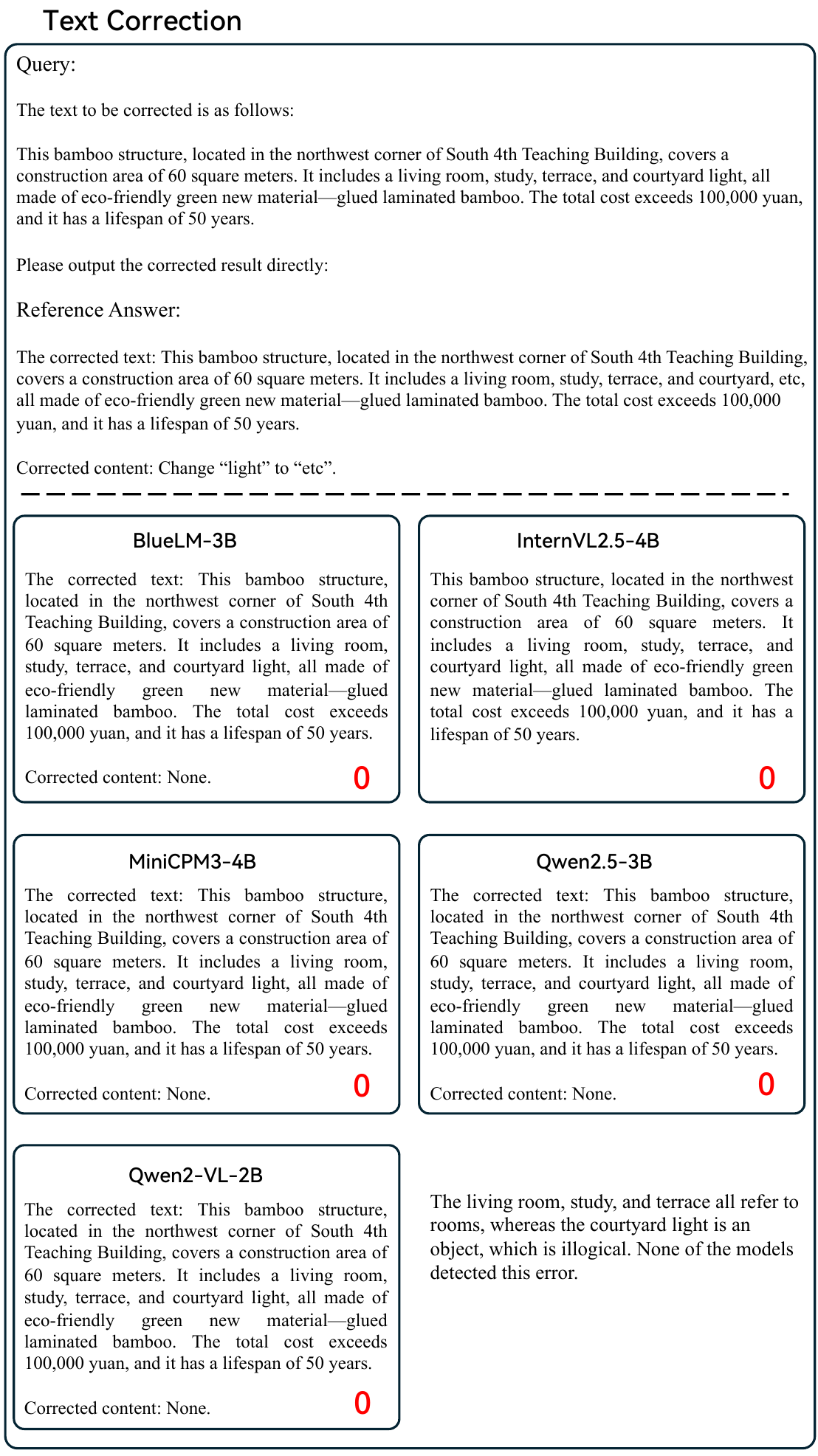}
    \caption{Text correction case (en).}
    \label{fig:example2_en}
\end{figure}

\begin{figure}[htbp]
    \centering
    \includegraphics[width=0.85\linewidth]{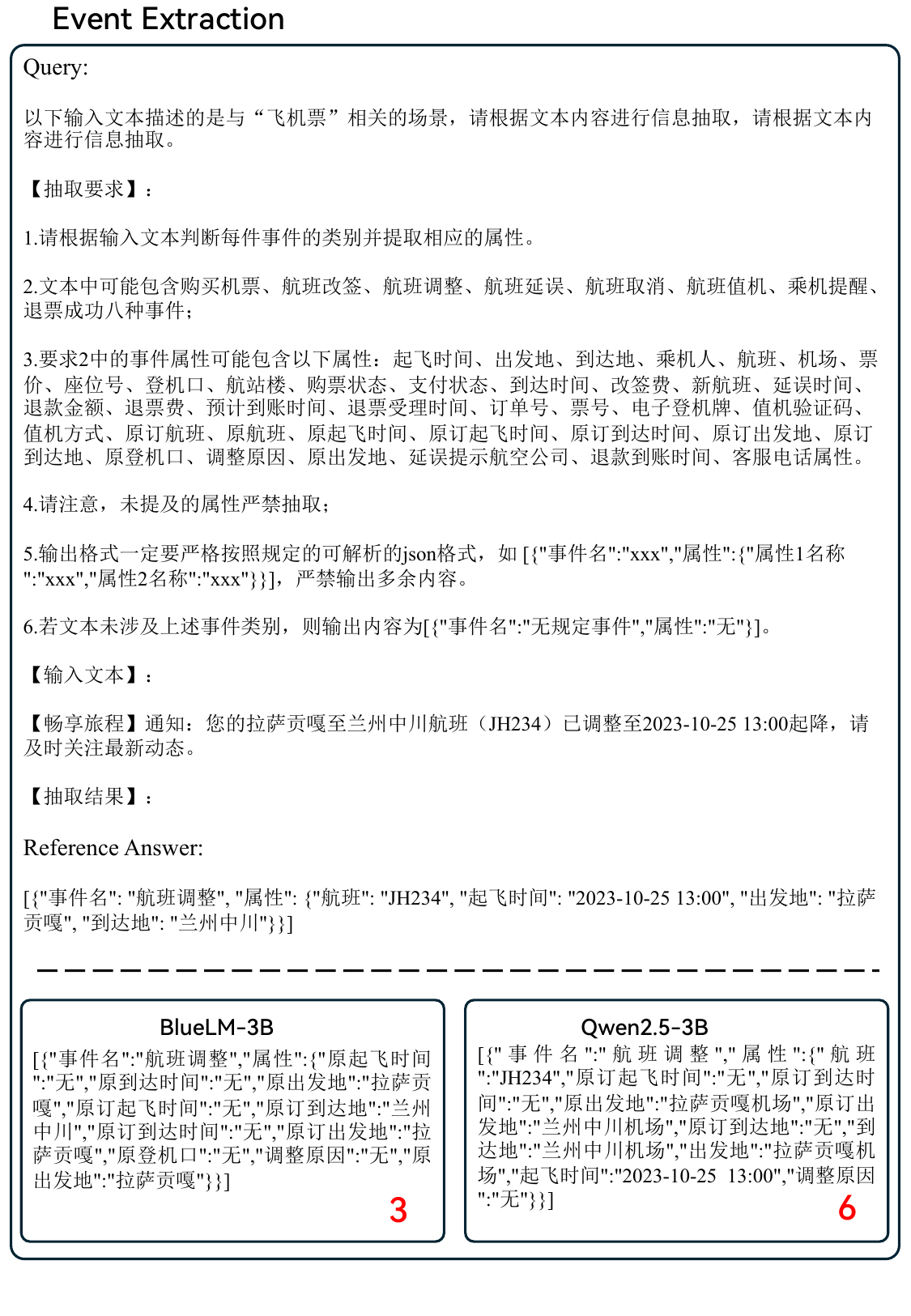}
    \caption{Event extraction case (zh).}
    \label{fig:example3}
\end{figure}

\begin{figure}[htbp]
    \centering
    \includegraphics[width=0.85\linewidth]{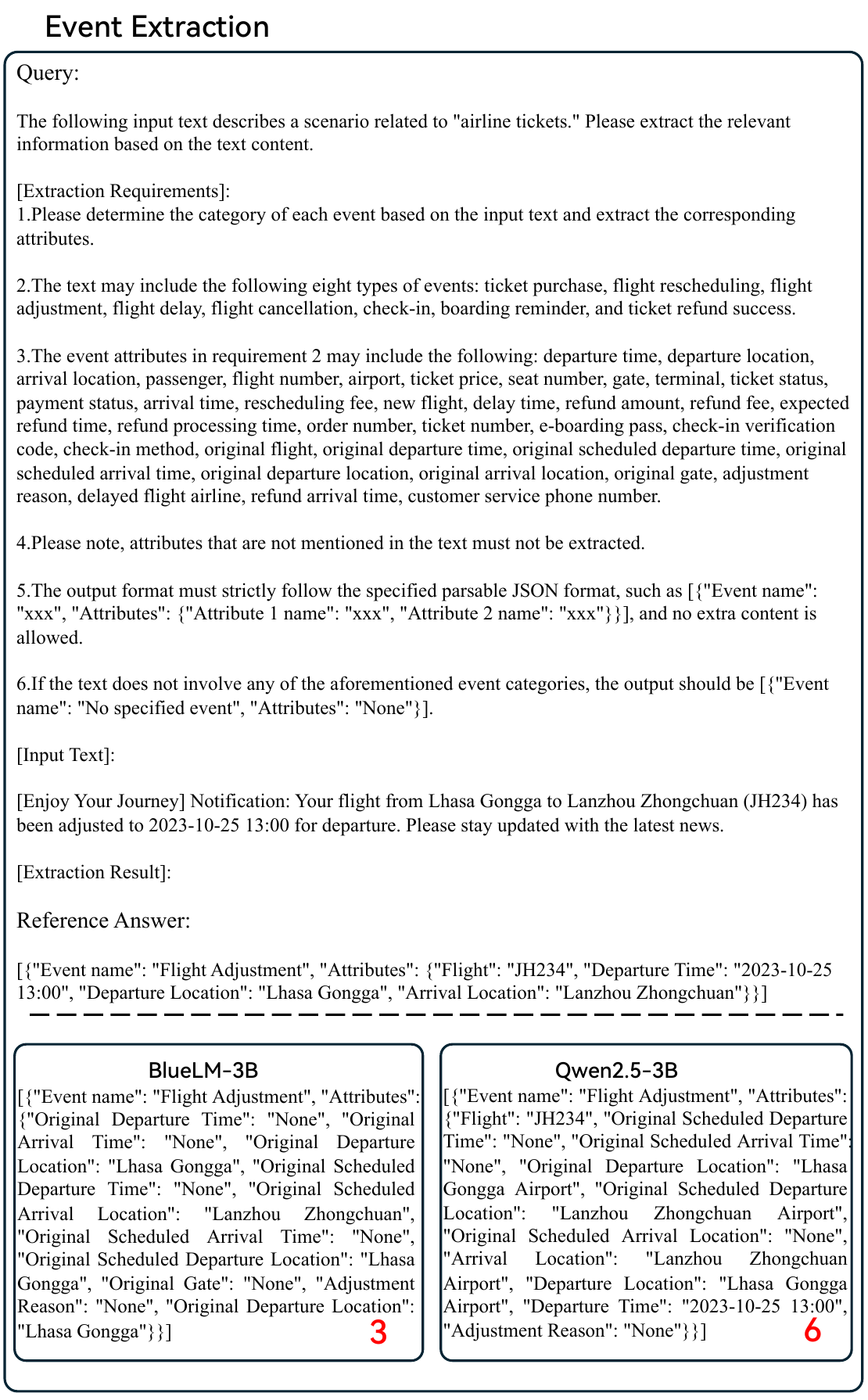}
    \caption{Event extraction case (en).}
    \label{fig:example3_en}
\end{figure}

\begin{figure}[htbp]
    \centering
    \includegraphics[width=0.85\linewidth]{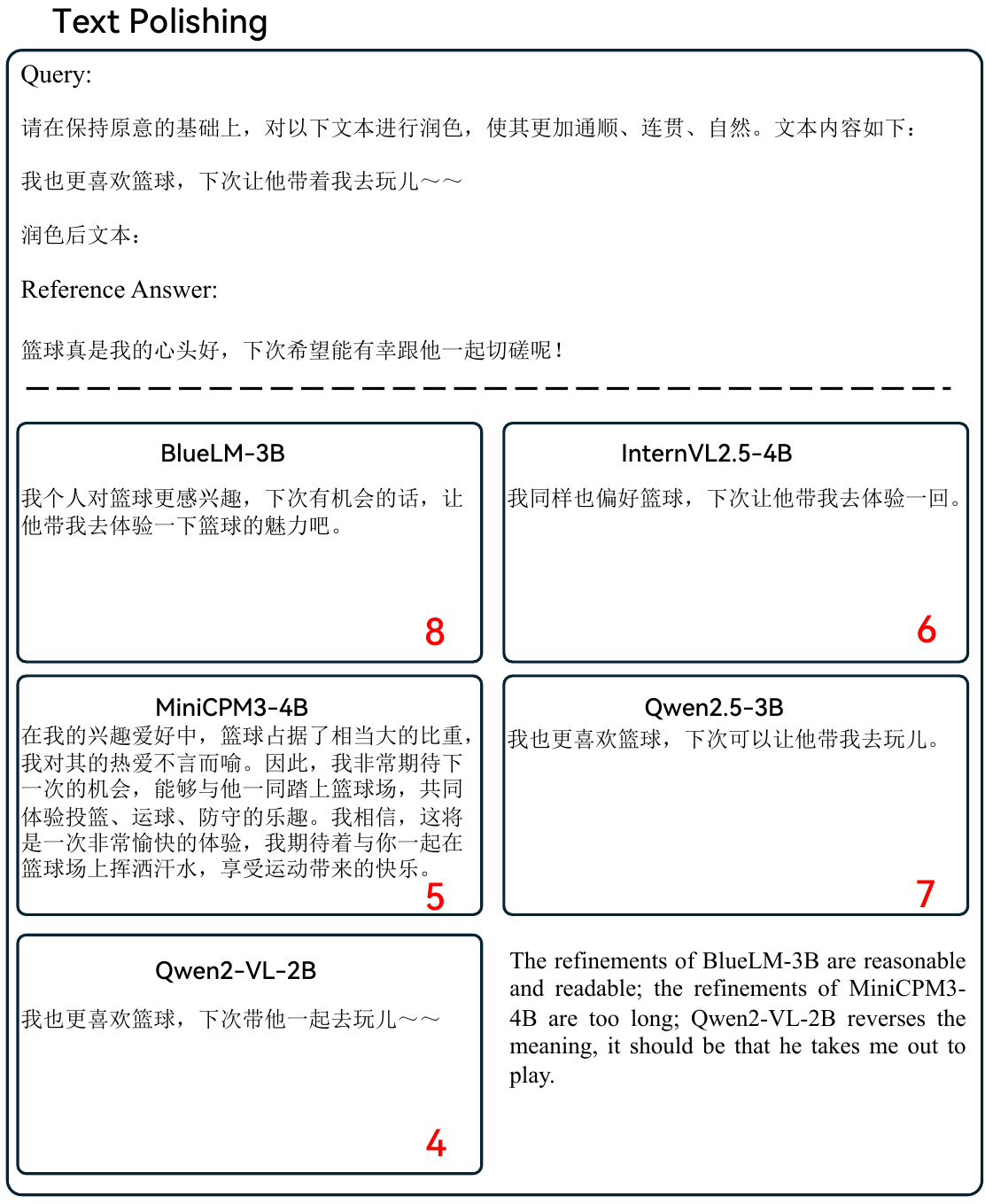}
    \caption{Text polishing case (zh).}
    \label{fig:example4}
\end{figure}

\begin{figure}[htbp]
    \centering
    \includegraphics[width=0.85\linewidth]{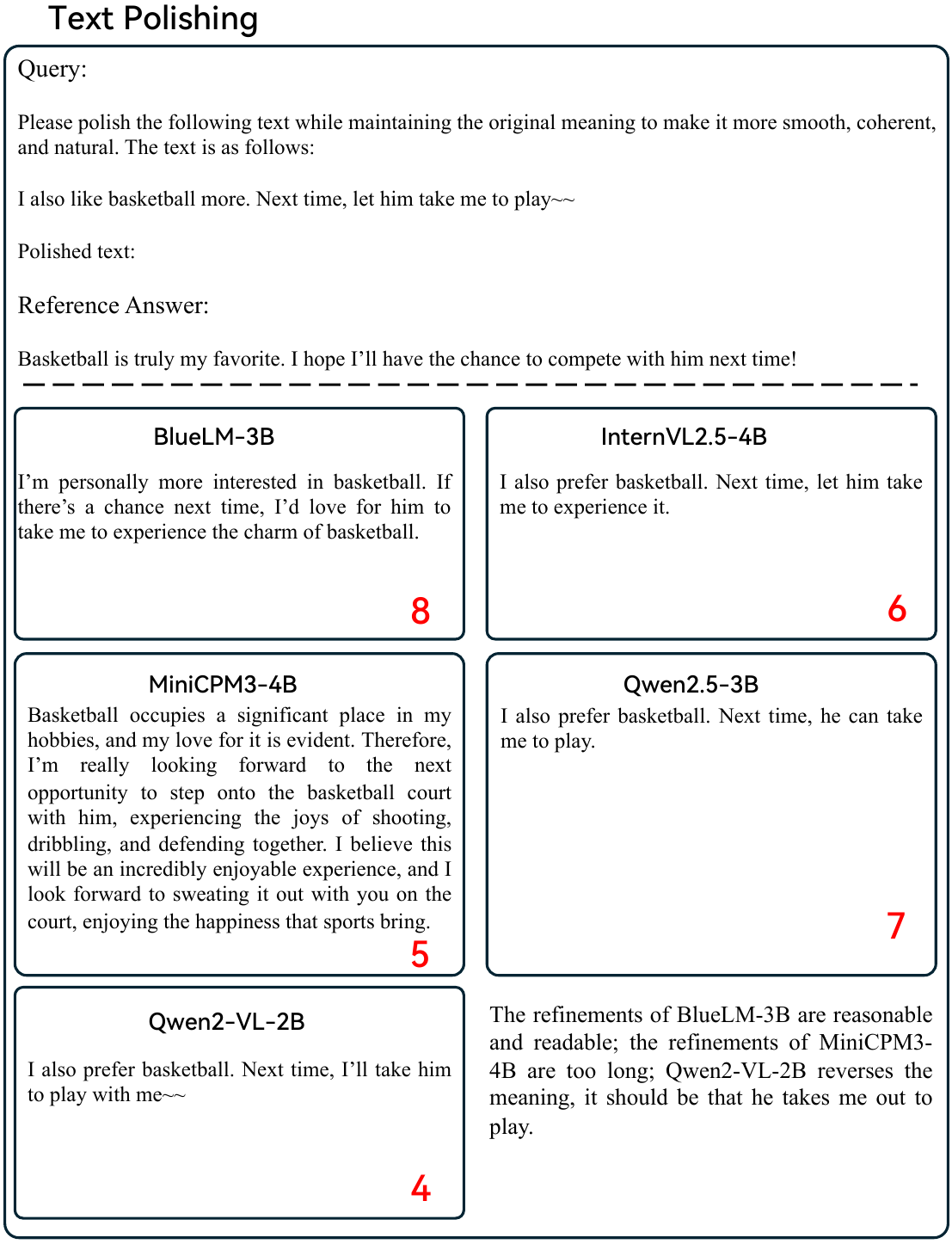}
    \caption{Text polishing case (en).}
    \label{fig:example4_en}
\end{figure}

\clearpage

\subsection{English Translation of Pictures in the Paper}

\begin{figure}[h]
    \centering
    \includegraphics[width=\linewidth]{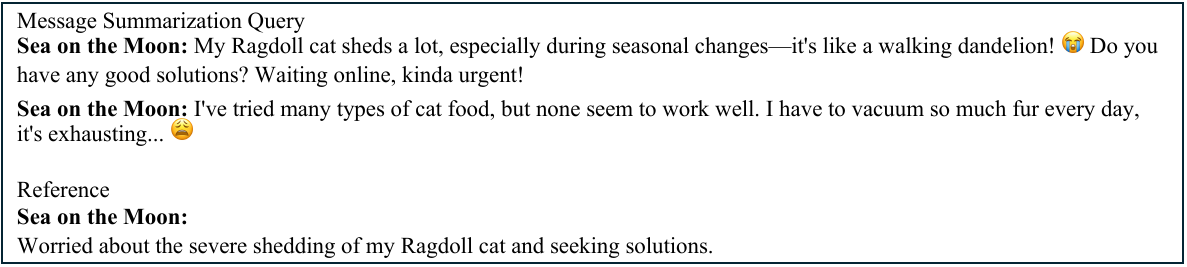}
    \caption{Translated example of the Message Summarization task in SmartBench.}
    \label{fig:en_case_example}
\end{figure}

\begin{figure}[h]
    \centering
    \includegraphics[width=\linewidth]{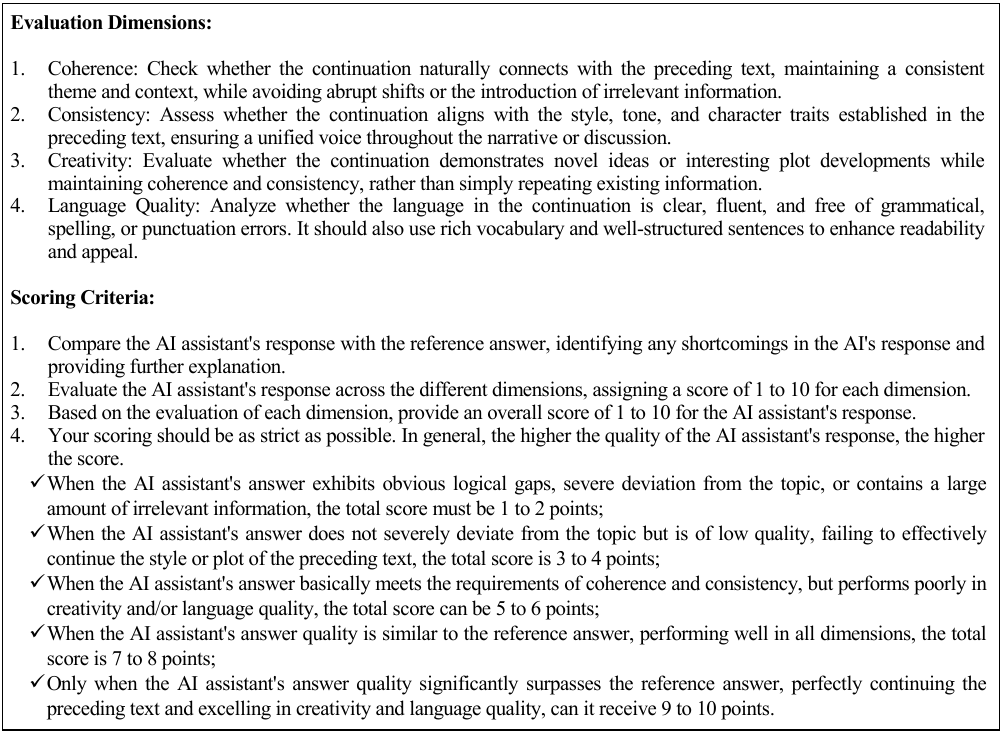}
    \caption{Evaluation Dimension \& Scoring Standard (in English) for the text continuation task.}
    \label{fig:en_eval_prompt}
\end{figure}

\end{document}